\documentclass[11pt]{article}
\usepackage{ifpdf}
\pdfoutput=1
\usepackage{geometry} 
\usepackage[parfill]{parskip}    % Activate to begin paragraphs with an empty line
\ifpdf
\usepackage[pdftex]{graphicx}
\else
\usepackage[dvips]{graphicx}
\fi
\usepackage{amsmath, amsthm}
\usepackage{amsfonts}
\usepackage{amssymb}
\usepackage{epstopdf}
\usepackage{subfigure}
\usepackage{fancyhdr}
\usepackage{algorithms/algorithmic}
\usepackage{algorithms/algorithm}
\usepackage{array}
\usepackage{url}
\usepackage{comment}

\newtheorem{thm}{Theorem}[section]

\theoremstyle{definition}
\newtheorem{defn}[thm]{Definition}
\newtheorem{rem}[thm]{Remark}
\numberwithin{equation}{section}
\usepackage{color}

\ifpdf
\usepackage[pdftex]{hyperref}
\else
\usepackage[dvips]{hyperref}
\fi

\newcommand{\vect}[1]{\boldsymbol{#1}}
\newcommand{\matr}[1]{\boldsymbol{#1}}

\newcommand{\mvn}[3]{\mathcal{N}\left( \vect{#1} \mid \vect{#2}, \matr{#3} \right)} % N(x | mu, sigma^2)
\newcommand{\svn}[3]{\mathcal{N}\left( #1 \mid #2, #3 \right)}

\newcommand{\deriv}[1]{\frac{\partial}{\partial #1}}

\title{PMOG: The projected mixture of Gaussians model with application to blind source separation}

\author{Gautam V. Pendse\thanks{To whom correspondence should be addressed. e-mail: gpendse@mclean.harvard.edu}$^{\,\,\, 1}$  \\
\mbox{}\\ \\ \\ \\ 
$^1$ P.A.I.N Group, Imaging and Analysis Group (IMAG), Harvard Medical School}
\date{July 22, 2010}
\begin{document}

\maketitle

\newpage 

\section{Abstract}
We extend the mixtures of Gaussians (MOG) model to the projected mixture of Gaussians (PMOG) model. In the PMOG model, we assume that $q$ dimensional input data points $\mathbf{z}_i$ are projected by a $q$ dimensional vector $\mathbf{w}$ into 1-D variables $u_i$. The projected variables $u_i$ are assumed to follow a 1-D MOG model. In the PMOG model, we maximize the likelihood of observing $u_i$ to find both the model parameters for the 1-D MOG as well as the projection vector $\mathbf{w}$. First, we derive an EM algorithm for estimating the PMOG model. Next, we show how the PMOG model can be applied to the problem of blind source separation (BSS). In contrast to conventional BSS where an objective function based on an approximation to differential entropy is minimized, PMOG based BSS simply minimizes the differential entropy of projected sources by fitting a flexible MOG model in the projected 1-D space while simultaneously optimizing the projection vector $\mathbf{w}$. The advantage of PMOG over conventional BSS algorithms is the more flexible fitting of non-Gaussian source densities without assuming near-Gaussianity (as in conventional BSS) and still retaining computational feasibility.

\section{Introduction}

The mixture of Gaussians (MOG) is a flexible model with application to many real world problems. The adjustable parameters in a 1-D MOG model include the number of component Gaussian distributions, the mean and variance of each component distribution and their mixing fractions. Given 1-D data points $u_i$, these distributional parameters can be efficiently estimated by maximum likelihood (ML) using the expectation maximization (EM) algorithm \cite{Dempster:1977}. Now consider the following situation:
\begin{itemize}
\item Data points $u_i$ are not given directly but suppose that we are given vectors $\vect{z_i}$. Next, 1-D scalar variables $u_i$ are generated by $u_i = \vect{w}^T \vect{z_i}$ where $\vect{w}$ is an unknown projection vector. 
\item Suppose that projected variables $u_i$ follow a 1-D MOG model which we will refer to as a projected mixture of Gaussians (PMOG) model in this work. 
\end{itemize}
Can we estimate the PMOG distributional parameters as well as the projection vector $\vect{w}$ using ML? In particular, can we derive an EM algorithm for this PMOG model similar to the standard EM algorithm for the conventional MOG model? 

While estimating the PMOG model is an interesting problem in its own right, we will show that it is also closely related to the problem of estimating the differential entropy of a random variable. Given this fact, we will show that the PMOG model can be applied to the problem of linear blind source separation (BSS) and linear independent component analysis (ICA). We use the term ICA to refer to "square mixing" where there are equal number of latent sources and mixed signals and BSS to refer to "non-square and noisy" mixing where there are more mixtures corrupted with additive Gaussian noise than latent sources. From this point of view, ICA is a special "noise free" case of BSS. 

BSS or ICA is a well studied problem and we refer the reader to \cite{Comon:1994, Hyvarinen:book, ICALAB_BOOK:2003} for a detailed overview of relevant work. A central point in BSS is choosing the "measuring function" for differential entropy of a random variable.
One of the most widely used algorithm for BSS is the FastICA (FICA) algorithm \cite{FPICA:1999}. The FICA algorithm works by optimizing the differential entropy based contrast functions developed in the seminal work by Hyvarinen et al.\ \cite{negentropy:1998}. These contrast functions are approximations to the differential entropy of a random variable $x$ under the following conditions:
\begin{itemize}
\item Expected values of certain functions $G_i(x)$ are given. The density of $x$ is estimated to be the maximum entropy distribution (MED) $f^{med}(x)$ subject to the these constraints.
\item Most importantly, the assumed MED $f^{med}(x)$ is, in the words of Hyvarinen et al.\ \cite{negentropy:1998} "not very far from a Gaussian distribution". Let us denote this simplified form of $f^{med}(x)$ by $f^{med}_{Gaussian}(x)$.
\end{itemize}
Hyvarinen et al.\ calculated expressions for the differential entropy using the distribution $f^{med}_{Gaussian}(x)$ given flexible user defined functions $G_i(x)$. A key question that arises is: are these approximations to differential entropy and BSS solutions based on $f^{med}_{Gaussian}(x)$ adequate when the true density and hence $f^{med}(x)$ is not "near Gaussian"? 

In another seminal paper, Attias et al.\ \cite{Attias:1999} developed a general solution to the BSS problem where the latent source density was modeled as a "factorial MOG" density. This significant advance removed the "near Gaussianity" assumption on the latent source densities. In addition, \cite{Attias:1999} developed an EM algorithm for the ML solution of the mixing parameters and MOG source parameters in BSS followed by a posterior mean or maximum aposteriori (MAP) estimate of the sources. However, this algorithm becomes computationally intractable for $> 13$ sources. Moreover, the solution by Attias et al.\ assumes "exact independence" between the sources i.e., it does not allow any partial dependence between sources. Can we develop a solution to the BSS problem that retains the flexible latent density modeling of  \cite{Attias:1999}, retains computational tractability and can be applied under partial dependence between sources? 

In this work, we develop the PMOG model and then apply it to the problem of BSS to address these questions:
\begin{enumerate}
\item We describe the PMOG model and derive an EM algorithm for estimating its parameters in \textbf{section \ref{pmog_model}}.
\item We show how PMOG can be applied to solving a BSS problem including cases where partial dependence between sources is allowed in \textbf{sections \ref{bss_approaches}, \ref{pmog_based_bss}}.

\end{enumerate}

\section{Notation}
\begin{itemize}
\item Scalars will be denoted in a non-bold font (e.g. $\gamma, \theta, L$) possibly with subscripts (e.g. $\pi_k$, $\mu_k$). We will use bold face lower case letters possibly with subscripts to denote vectors (e.g. $\vect{\mu}, \vect{x}, \vect{z_1}$) and bold face upper case letters possibly with subscripts to denote matrices (e.g. $\matr{A}, \matr{\Sigma}, \matr{B_1}$). The transpose of a matrix $\matr{A}$ will be denoted by $\matr{A}^T$ and its inverse will be denoted by $\matr{A}^{-1}$. We will denote the $p \times p$ identity matrix by $\matr{I}_p$. A vector or matrix of all zeros will be denoted by a bold face zero $\mathbf{0}$ whose size should be clear from context.  

\item The $j$th component of vector $\vect{x_i}$ will be denoted by $x_{ij}$ whereas the $j$th component of vector $\vect{x}$ will be denoted by $x_{j}$. The element $(i,j)$ of matrix $\matr{A}$ will be denoted by $A(i,j)$. Estimates of variables will be denoted by putting a hat on top of the variable symbol. For example, an estimate of $\sigma^2$ will be denoted by $\hat{\sigma}^2$. The 2-norm of a $p \times 1$ vector $\vect{x}$ will be denoted by $|| \vect{x} ||_2 = +\sqrt{ \sum_{i = 1}^p x_i^2 }$.

\item If $\vect{x}$ is a random vector with a multivariate Normal distribution with mean $\vect{\mu}$ and covariance $\matr{\Sigma}$ then we will denote this distribution by $\mvn{x}{\mu}{\Sigma}$. Similarly, if $u$ is a scalar random variable with a Normal distribution with mean $\mu_k$ and variance $\sigma_k^2$ then we will denote this distribution by $\svn{u}{\mu_k}{\sigma_k^2}$. We will use $U(a,b)$ to denote the uniform random distribution on $(a,b)$. Unless otherwise stated all logarithms are natural logarithms (i.e., $\log$ means $\log_{e}$). Probability density functions will be denoted by capital letters with the vector of arguments in parenthesis such as $P(\vect{x})$ or $Q(\vect{x})$. When necessary we will also indicate the dependence of a probability density on a vector of parameters $\vect{\theta}$ by using the notation $P(\vect{x} \mid \vect{\theta})$. The expected value of a function $f(\vect{x})$ with respect to the density $P(\vect{x})$ will be denoted by $E_x[f(\vect{x})]$ or simply by $E[f(\vect{x})]$.
\end{itemize}

\section{Projected mixture of Gaussians (PMOG) model}\label{pmog_model}
In this section, we try to answer the following questions:
\begin{itemize}
\item What is the PMOG model? What is given and what needs to be estimated?
\item Can we derive an EM algorithm for estimating the PMOG model?
\item How are the MOG parameters and projection vector estimated in the M-step?
\item What other precautions need to be taken to make the monotonic likelihood improvement property of EM hold true?
\end{itemize}

\subsection{What is the PMOG model?}

\begin{figure}[htbp]
\begin{center}
\includegraphics[width = 6.5in] {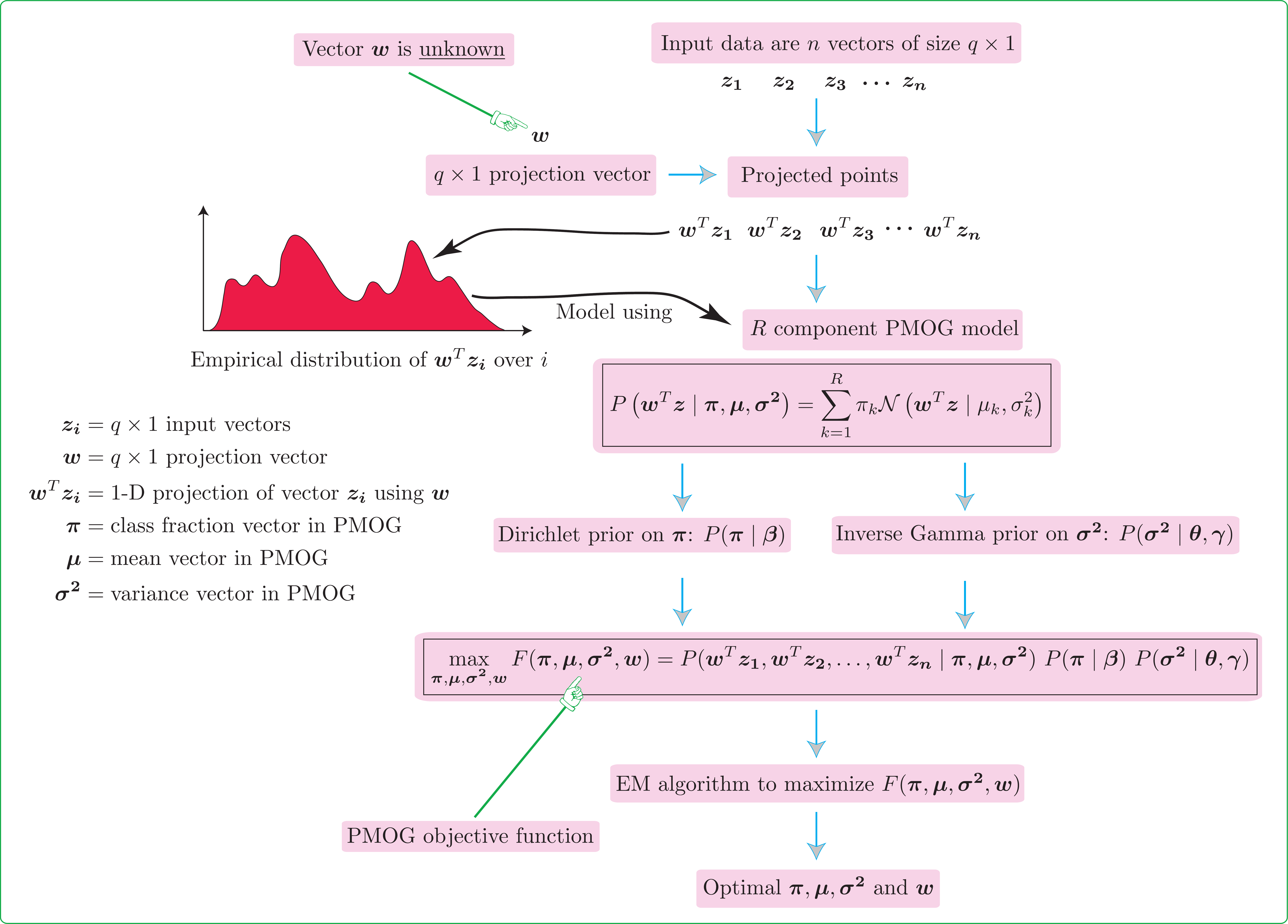}
\caption{Pictorial depiction of the PMOG model. Input vectors $\vect{z_i}$ are projected into 1-D variables using projection vector $\vect{w}$. The 1-D empirical density of projected points $\vect{w}^T \vect{z_i}$ is modeled using a MOG density. In contrast to conventional MOG, the PMOG model requires estimation of both the MOG parameters $\vect{\pi}, \vect{\mu}, \vect{\sigma^2}$ as well as the projection vector $\vect{w}$. The overall objective function $F$ also includes the influence of priors for $\vect{\pi}$ and $\vect{\sigma^2}$. This is done to prevent the collapse of a component density of PMOG onto a single point.}
\label{figure1}
\end{center}
\end{figure}

Suppose we are given $n$ vectors $\vect{z_1},\vect{z_2},\ldots, \vect{z_n}$ where each $\vect{z_i}$ is a $q \times 1$ vector. Suppose $\matr{Z}$ is a $q \times n$ matrix formed by assembling these vectors into a matrix i.e., 
\begin{equation}\label{eq1}
\matr{Z} = [\vect{z_1},\vect{z_2},\ldots, \vect{z_n}]
\end{equation}
We can think of these vectors as $n$ realizations of a random vector $\vect{z}$. Suppose $\vect{w}$ is an unknown $q \times 1$ vector that defines new "projected" scalars $u_1,u_2,\ldots,u_n$ such that 
\begin{equation}\label{eq2}
u_i = \vect{w}^T \vect{z_i} = \vect{z_i}^T \vect{w}
\end{equation}
Again, these $n$ scalars can be thought of as $n$ realizations of the random variable $u = \vect{w}^T \vect{z}$. Suppose we are also given a $q \times L$ matrix $\matr{G}$ of full column rank $L$ with $q > L$. It is given that the unknown vector $\vect{w}$ satisfies the constraints
\begin{align}\label{eq3}
\vect{w}^T \vect{w} &= 1\\
\matr{G}^T \vect{w} &=  \matr{0}
\end{align}
It is assumed that the density of random variable $u$ is a mixture of $R$ Gaussians i.e.,
\begin{equation}\label{eq4}
P(u \mid \vect{\pi}, \vect{\mu}, \vect{\sigma^2}) = \sum_{k = 1}^R \pi_k \, \svn{u}{\mu_k}{\sigma_k^2}
\end{equation}
In the above equation, $\vect{\pi}$ is a $R \times 1$ vector  $[\pi_1,\pi_2,\ldots,\pi_R]^T$. Similarly $\vect{\mu} = [\mu_1,\mu_2,\ldots,\mu_R]^T$ and $\vect{\sigma^2} = [\sigma^2_1,\sigma^2_2,\ldots,\sigma^2_R]^T$ are also $R \times 1$ vectors. The class fractions $\pi_k$ satisfy:
\begin{gather}\label{eq4a}
\sum_{k = 1}^R \pi_k = 1 \\
0 \le \pi_k \le 1
\end{gather}
Equation \ref{eq4} can be equivalently written in terms of $\vect{z}$ as follows:
\begin{equation}\label{eq5}
\boxed{
P( \vect{w}^T \vect{z} \mid \vect{\pi}, \vect{\mu}, \vect{\sigma^2}) = \sum_{k = 1}^R \pi_k \, \svn{ \vect{w}^T \vect{z} }{\mu_k}{\sigma_k^2}
}
\end{equation}
We will call the model \ref{eq5} a projected mixture of Gaussians or PMOG model. A pictorial description of the PMOG model is given in Fig. \ref{figure1}.

\subsection{Estimating the PMOG model}

Assuming that $\vect{w}^T \vect{z_i}$ for $i = 1,2,\ldots,n$ are $n$ independent realizations of $\vect{w}^T \vect{z}$, we can write their joint density as
\begin{equation}\label{eq6}
P(\vect{w}^T \vect{z_1}, \vect{w}^T \vect{z_2},\ldots,\vect{w}^T \vect{z_n} \mid \vect{\pi}, \vect{\mu}, \vect{\sigma^2}) = \prod_{i = 1}^n P(\vect{w}^T \vect{z_i} \mid \vect{\pi}, \vect{\mu}, \vect{\sigma^2})
\end{equation} 
To simplify notation, we will denote the left hand size of \ref{eq6} by $P(\matr{Z}^T \vect{w} \mid \vect{\pi}, \vect{\mu}, \vect{\sigma^2})$. With this notation, we can write:
\begin{equation}\label{eq7}
P(\matr{Z}^T \vect{w} \mid \vect{\pi}, \vect{\mu}, \vect{\sigma^2}) = \prod_{i = 1}^n P(\vect{w}^T \vect{z_i} \mid \vect{\pi}, \vect{\mu}, \vect{\sigma^2})
\end{equation}
The problem is to \textbf{maximize} $P(\matr{Z}^T \vect{w} \mid \vect{\pi}, \vect{\mu}, \vect{\sigma^2})$ (equation \ref{eq7}) w.r.t $\vect{\pi}, \vect{\mu}, \vect{\sigma^2}$ and $\vect{w}$. For a given $\vect{w}$ this problem is equivalent to the standard mixture of Gaussians (MOG) estimation problem and can be handled efficiently by the expectation maximization (EM) algorithm \cite{Dempster:1977}. In the PMOG model, the novelty is to allow $\vect{w}$ to be an unknown that is estimated along with MOG parameters $\vect{\pi}, \vect{\mu}, \vect{\sigma^2}$.
Next we introduce priors on $\vect{\pi}$ and $\vect{\sigma^2}$. The purpose of introducing priors on $\vect{\pi}$ and $\vect{\sigma^2}$ is simply to prevent the collapse of a Gaussian component from PMOG onto a single data point. Suppose we assume a Dirichlet prior on the vector $\vect{\pi}$. This prior is described by a $R \times 1$ parameter vector $\vect{\beta}$ with elements $\beta_1, \beta_2,\ldots, \beta_R$.
\begin{equation}\label{eq8}
P(\vect{\pi} \mid \vect{\beta}) \propto \prod_{k = 1}^R \pi_k^{\beta_k - 1}
\end{equation}
Similarly we assume a product of inverse Gamma prior on $\vect{\sigma^2}$. Suppose $\vect{\theta}$ is a vector with elements $\theta_1,\theta_2,\ldots,\theta_R$ and similarly $\vect{\gamma}$ is a vector with elements $\gamma_1,\gamma_2,\ldots,\gamma_R$ then
\begin{equation}\label{eq9}
P(\vect{\sigma^2} \mid \vect{\theta}, \vect{\gamma}) = \prod_{k = 1}^R P(\sigma_k^2 \mid \theta_k, \gamma_k)  \propto \prod_{k = 1}^R \left(\sigma_k^2\right)^{-(\theta_k + 1)} \, \exp\left( {-\dfrac{1}{\gamma_k \, \sigma_k^2}} \right)
\end{equation}
We use an improper prior for $\vect{\mu}$, $P(\vect{\mu}) = 1$. With this choice of priors the posterior distribution can be written as:
\begin{equation}\label{eq10}
P(\vect{\pi}, \vect{\mu}, \vect{\sigma^2} \mid \matr{Z}^T \vect{w}) = \frac{P(\matr{Z}^T \vect{w} \mid \vect{\pi}, \vect{\mu}, \vect{\sigma^2}) \,\, P(\vect{\pi} \mid \vect{\beta}) \,\, P(\vect{\sigma^2} \mid \vect{\theta}, \vect{\gamma}) }{ P(\matr{Z}^T \vect{w})}
\end{equation}
Maximization of this posterior density w.r.t both $\vect{\pi}, \vect{\mu}, \vect{\sigma^2}$ and $\vect{w}$ is difficult because of the presence of $\vect{w}$ in the denominator. We could however, maximize the posterior density w.r.t $\vect{\pi}, \vect{\mu}, \vect{\sigma^2}$ only and maximize the likelihood term $P(\matr{Z}^T \vect{w} \mid \vect{\pi}, \vect{\mu}, \vect{\sigma^2})$ w.r.t $\vect{w}$. This is equivalent to solving the problem:
\begin{equation}\label{eq11}
\boxed{
\max_{\vect{\pi}, \vect{\mu}, \vect{\sigma^2}, \vect{w}} F(\vect{\pi}, \vect{\mu}, \vect{\sigma^2}, \vect{w}) = P(\matr{Z}^T \vect{w} \mid \vect{\pi}, \vect{\mu}, \vect{\sigma^2}) \,\, P(\vect{\pi} \mid \vect{\beta}) \,\, P(\vect{\sigma^2} \mid \vect{\theta}, \vect{\gamma}) 
}
\end{equation}

\subsection{EM algorithm for maximizing $F$}

The main development in this subsection is an EM algorithm for maximizing $F$. It is much easier to deal with the logarithm of $F$ rather than $F$ itself. Now 
\begin{equation}\label{eq12}
\log F(\vect{\pi}, \vect{\mu}, \vect{\sigma^2}, \vect{w}) = \log P(\matr{Z}^T \vect{w} \mid \vect{\pi}, \vect{\mu}, \vect{\sigma^2}) + \log P(\vect{\pi} \mid \vect{\beta}) + \log P(\vect{\sigma^2} \mid \vect{\theta}, \vect{\gamma}) 
\end{equation}
Substituting \ref{eq7}, \ref{eq8} and \ref{eq9} into \ref{eq12} we get:
\begin{align}\label{eq13}
& \log F(\vect{\pi}, \vect{\mu}, \vect{\sigma^2}, \vect{w}) \\ \nonumber
& \propto  \log \left( \prod_{i = 1}^n P(\vect{w}^T \vect{z_i} \mid \vect{\pi}, \vect{\mu}, \vect{\sigma^2}) \right) + \log \left( \prod_{k = 1}^R \pi_k^{\beta_k - 1} \right) + \log \left( \prod_{k = 1}^R \left(\sigma_k^2\right)^{-(\theta_k + 1)} \, \exp\left( {-\dfrac{1}{\gamma_k \, \sigma_k^2}} \right) \right)
\end{align}
which can be simplified to:
\begin{align}\label{eq14}
& \log F(\vect{\pi}, \vect{\mu}, \vect{\sigma^2}, \vect{w}) \\ \nonumber
& \propto  \sum_{i = 1}^n \log P(\vect{w}^T \vect{z_i} \mid \vect{\pi}, \vect{\mu}, \vect{\sigma^2}) \,+\, \sum_{k = 1}^R (\beta_k - 1) \log \pi_k \,+\, \sum_{k = 1}^R  \left( -(\theta_k + 1) \log \sigma_k^2 - \dfrac{1}{\gamma_k \, \sigma_k^2} \right)
\end{align}
In the above equation, we have ignored the constants of the prior densities on $\vect{\pi}$ and $\vect{\sigma^2}$ since they do not depend on the unknown parameters. Thus the objective function to be maximized can be written as:
\begin{equation}\label{eq15}
H(\vect{\pi}, \vect{\mu}, \vect{\sigma^2}, \vect{w}) = H_1( \vect{\pi}, \vect{\mu}, \vect{\sigma^2}, \vect{w} ) + H_2(\vect{\pi}, \vect{\sigma^2})
\end{equation}
where
\begin{equation}\label{eq16}
 H_1( \vect{\pi}, \vect{\mu}, \vect{\sigma^2}, \vect{w} )  = \sum_{i = 1}^n \log P(\vect{w}^T \vect{z_i} \mid \vect{\pi}, \vect{\mu}, \vect{\sigma^2})
\end{equation}
and
\begin{equation}\label{eq17}
H_2(\vect{\pi}, \vect{\sigma^2}) = \sum_{k = 1}^R (\beta_k - 1) \log \pi_k \,+\, \sum_{k = 1}^R  \left( -(\theta_k + 1) \log \sigma_k^2 - \dfrac{1}{\gamma_k \, \sigma_k^2} \right)
\end{equation}
Substituting \ref{eq5} into \ref{eq16} we get:
\begin{equation}\label{eq18}
H_1( \vect{\pi}, \vect{\mu}, \vect{\sigma^2}, \vect{w} )  = \sum_{i = 1}^n \log \sum_{k = 1}^R \pi_k \, \svn{ \vect{w}^T \vect{z_i} }{\mu_k}{\sigma_k^2}
\end{equation}
Introducing auxillary variables $\alpha_{ki}$ such that:
\begin{gather}
 \sum_{k = 1}^R \alpha_{ki} = 1 \\
 0 \le \alpha_{ki} \le 1
 \end{gather}
we can write
\begin{equation}\label{eq19}
H_1( \vect{\pi}, \vect{\mu}, \vect{\sigma^2}, \vect{w} )  = \sum_{i = 1}^n \log \sum_{k = 1}^R \left( \frac{ \pi_k \, \svn{ \vect{w}^T \vect{z_i} }{\mu_k}{\sigma_k^2} }{\alpha_{ki}} \right) \alpha_{ki}
\end{equation}
%By Jensen's inequality, we can write
By using the concavity of the $\log$ function, we can write:
\begin{equation}\label{eq19a}
H_1( \vect{\pi}, \vect{\mu}, \vect{\sigma^2}, \vect{w} )  \ge Q( \vect{\pi}, \vect{\mu}, \vect{\sigma^2}, \vect{w}, \matr{\alpha} ) = \sum_{i = 1}^n \sum_{k = 1}^R \alpha_{ki} \log \left( \frac{ \pi_k \, \svn{ \vect{w}^T \vect{z_i} }{\mu_k}{\sigma_k^2} }{\alpha_{ki}} \right) 
\end{equation}
From \ref{eq15} we see that
\begin{equation}\label{eq20}
H(\vect{\pi}, \vect{\mu}, \vect{\sigma^2}, \vect{w}) = H_1( \vect{\pi}, \vect{\mu}, \vect{\sigma^2}, \vect{w} ) + H_2(\vect{\pi}, \vect{\sigma^2}) \ge Q( \vect{\pi}, \vect{\mu}, \vect{\sigma^2}, \vect{w}, \matr{\alpha} ) + H_2(\vect{\pi}, \vect{\sigma^2})
\end{equation}
The rightmost inequality becomes an equality when we choose: 
\begin{equation}\label{eq21}
\alpha_{ki} = \frac{ \pi_k \, \svn{\vect{w}^T \vect{z_i}}{\mu_k}{\sigma_k^2} }{ \sum_{k=1}^R \pi_k \, \svn{\vect{w}^T \vect{z_i}}{\mu_k}{\sigma_k^2} }
\end{equation}
In the context of EM, the $\alpha_{ki}$ are also known as responsibilities.
Suppose $\vect{\pi}^{(t)}, \vect{\mu}^{(t)}, \vect{\sigma^2}^{(t)}, \vect{w}^{(t)}$ are the parameter values at iteration $t$ and $\matr{\alpha}^{(t)}$ be the corresponding responsibilities computed from \ref{eq21}. Then we have
\begin{align}\label{eq22a}
& H \left( \vect{\pi}^{(t+1)}, \vect{\mu}^{(t+1)}, \vect{\sigma^2}^{(t+1)}, \vect{w}^{(t+1)} \right) \\ \nonumber
& = Q\left( \vect{\pi}^{(t+1)}, \vect{\mu}^{(t+1)}, \vect{\sigma^2}^{(t+1)}, \vect{w}^{(t+1)}, \matr{\alpha}^{(t+1)} \right) + H_2\left( \vect{\pi}^{(t+1)}, \vect{\sigma^2}^{(t+1)} \right) && \text{(by \ref{eq20} and \ref{eq21})} \\ \nonumber
& \ge Q\left( \vect{\pi}^{(t+1)}, \vect{\mu}^{(t+1)}, \vect{\sigma^2}^{(t+1)}, \vect{w}^{(t+1)}, \matr{\alpha}^{(t)} \right) + H_2\left( \vect{\pi}^{(t+1)}, \vect{\sigma^2}^{(t+1)} \right)  && \text{(by \ref{eq20} and \ref{eq21})}
\end{align}
Now suppose, given $\vect{\pi}^{(t)}, \vect{\mu}^{(t)}, \vect{\sigma^2}^{(t)}, \vect{w}^{(t)}$ and $\vect{\alpha}^{(t)}$, we calculate $\vect{\pi}^{(t+1)}, \vect{\mu}^{(t+1)}, \vect{\sigma^2}^{(t+1)}, \vect{w}^{(t+1)}$ such that
\begin{align}\label{eq23a}
&Q\left( \vect{\pi}^{(t+1)}, \vect{\mu}^{(t+1)}, \vect{\sigma^2}^{(t+1)}, \vect{w}^{(t+1)}, \matr{\alpha}^{(t)} \right) + H_2\left( \vect{\pi}^{(t+1)}, \vect{\sigma^2}^{(t+1)} \right) \\ \nonumber
&\ge Q\left( \vect{\pi}^{(t)}, \vect{\mu}^{(t)}, \vect{\sigma^2}^{(t)}, \vect{w}^{(t)}, \matr{\alpha}^{(t)} \right) + H_2\left( \vect{\pi}^{(t)}, \vect{\sigma^2}^{(t)} \right)
\end{align}
From \ref{eq23a} and \ref{eq22a} we get:
\begin{align}\label{eq22}
& H \left( \vect{\pi}^{(t+1)}, \vect{\mu}^{(t+1)}, \vect{\sigma^2}^{(t+1)}, \vect{w}^{(t+1)} \right) \\ \nonumber
&\ge Q\left( \vect{\pi}^{(t)}, \vect{\mu}^{(t)}, \vect{\sigma^2}^{(t)}, \vect{w}^{(t)}, \matr{\alpha}^{(t)} \right) + H_2\left( \vect{\pi}^{(t)}, \vect{\sigma^2}^{(t)} \right) \\ \nonumber
&= H \left( \vect{\pi}^{(t)}, \vect{\mu}^{(t)}, \vect{\sigma^2}^{(t)}, \vect{w}^{(t)} \right) && \text{(by \ref{eq20} and \ref{eq21})}
\end{align}
This shows that the objective function $H$ increases monotonically from iteration $t$ to iteration $(t+1)$ and hence converges to a local maximum. This gives us the following EM algorithm for the PMOG model: 
\paragraph{Initialization}
Choose initial values of parameters either randomly or by other techniques such as clustering using $k$-means \cite{MacQueen:1967}. The output of this stage is the initial values $\vect{\pi}^{(0)}, \vect{\mu}^{(0)}, \vect{\sigma^2}^{(0)}, \vect{w}^{(0)}$. Following initialization, the E and M steps outlined below are performed in an alternating fashion until convergence.

\paragraph{E-step}
In the E-step, we calculate the responsibilities $\matr{\alpha}^{(t)}$ using the current estimate of parameters $\vect{\pi}^{(t)}, \vect{\mu}^{(t)}, \vect{\sigma^2}^{(t)}, \vect{w}^{(t)}$ and equation \ref{eq21}.

\paragraph{M-step}
In the M-step, we solve the problem:
\begin{equation}\label{eq23}
\vect{\pi}^{(t+1)}, \vect{\mu}^{(t+1)}, \vect{\sigma^2}^{(t+1)}, \vect{w}^{(t+1)} = \mbox{arg} \max_{\vect{\pi}, \vect{\mu}, \vect{\sigma^2}, \vect{w}} Q\left( \vect{\pi}, \vect{\mu}, \vect{\sigma^2}, \vect{w}, \matr{\alpha}^{(t)} \right) + H_2\left( \vect{\pi}, \vect{\sigma^2} \right)
\end{equation}

In ordinary MOG, the objective function in \ref{eq23} is a convex function of $\vect{\pi}, \vect{\mu}, \vect{\sigma^2}$ and a closed form solution exists for the M-step. If optimizing also w.r.t $\vect{w}$ then the objective function becomes non-convex and hence we need to explicitly impose the post optimization condition \ref{eq23a} for EM convergence. We use a relative error criterion to detect EM convergence. Our convergence criterion is:
\begin{gather}\label{conv_criterion}
\mbox{abs} \left[ H(\vect{\pi}^{(t+1)}, \vect{\mu}^{(t+1)}, \vect{\sigma^2}^{(t+1)}, \vect{w}^{(t+1)}) - H( \vect{\pi}^{(t)}, \vect{\mu}^{(t)}, \vect{\sigma^2}^{(t)}, \vect{w}^{(t)}) \right] \le \varepsilon^{*} \\
\varepsilon^{*} = \varepsilon_{rel} \, \left( \underset{t}{\mbox{mean}} \,\, \mbox{abs} \left[ H( \vect{\pi}^{(t)}, \vect{\mu}^{(t)}, \vect{\sigma^2}^{(t)}, \vect{w}^{(t)}) \right] \right)
\end{gather}
Here $\varepsilon_{rel}$ is a user specified relative error. We used $\varepsilon_{rel} = 10^{-5}$ in our experiments.

\subsection{Solving the M-step problem}
In this subsection, we discuss in detail the solution of problem \ref{eq23}. During the estimation of each variable $\vect{\pi}, \vect{\mu}, \vect{\sigma^2}$ and $\vect{w}$ we account for the relevant constraints using Lagrange multipliers and Karush-Kuhn-Tucker optimality conditions (see \cite{Nocedal:book} for details).

\paragraph{Estimating $\vect{\pi}$}
Differentiating the M-step objective function w.r.t $\pi_k$ and noting the constraint $\sum_{k = 1}^R \pi_k - 1 = 0$ the optimality condition is given by:
\begin{equation}\label{eq24}
\deriv{\pi_k} Q\left( \vect{\pi}, \vect{\mu}, \vect{\sigma^2}, \vect{w}, \matr{\alpha}^{(t)} \right) + \deriv{\pi_k} H_2\left( \vect{\pi}, \vect{\sigma^2} \right) - \lambda \deriv{\pi_k} ( \sum_{k = 1}^R \pi_k - 1 ) = 0
\end{equation}
Now,
\begin{align}\label{eq25}
\deriv{\pi_k} Q\left( \vect{\pi}, \vect{\mu}, \vect{\sigma^2}, \vect{w}, \matr{\alpha}^{(t)} \right) &= \frac{1}{\pi_k} \sum_{i = 1}^n \alpha^{(t)}_{ki} \\
\deriv{\pi_k} H_2\left( \vect{\pi}, \vect{\sigma^2} \right) &= \frac{1}{\pi_k} (\beta_k - 1)
\end{align}
From \ref{eq24} and \ref{eq25} we get:
\begin{equation}\label{eq26}
\frac{1}{\pi_k} \sum_{i = 1}^n \alpha^{(t)}_{ki} + \frac{1}{\pi_k} (\beta_k - 1) = \lambda \Longrightarrow \frac{1}{\lambda} \left( \sum_{i = 1}^n \alpha^{(t)}_{ki} +  (\beta_k - 1) \right) = \pi_k
\end{equation}
Imposing the constraint $\sum_{k = 1}^R \pi_k = 1$ and noting that $\sum_{k = 1}^R \alpha^{(t)}_{ki} = 1$ we get:
\begin{equation}\label{eq27}
\lambda = \sum_{k=1}^R \left( \sum_{i = 1}^n \alpha^{(t)}_{ki} +  (\beta_k - 1) \right) \Longrightarrow \lambda =  n +  \sum_{k=1}^R (\beta_k - 1) 
\end{equation}
Thus $\pi_k$ is given by:
\begin{equation}\label{eq28}
\pi_k = \frac{ \sum_{i = 1}^n \alpha^{(t)}_{ki} +  (\beta_k - 1)  }{n +  \sum_{k=1}^R (\beta_k - 1) }
\end{equation}
We have ignored the inequality constraints $0 \le \pi_k \le 1$ on $\pi_k$ during optimization. By choosing $\beta_k > 1$ we can ensure that these constraints are always satisfied and inactive and hence can be disregarded during optimization.

\paragraph{Estimating $\vect{\mu}$}
Differentiating the M-step objective w.r.t $\mu_k$ and noting that $H_2\left( \vect{\pi}, \vect{\sigma^2} \right)$ is not dependent on $\mu_k$, the optimality condition is given by:
\begin{align}\label{eq29}
\deriv{\mu_k} Q\left( \vect{\pi}, \vect{\mu}, \vect{\sigma^2}, \vect{w}, \matr{\alpha}^{(t)} \right) = 0
\end{align}
Upon substituting the derivative, we get
\begin{equation}\label{eq30}
\sum_{i = 1}^n \alpha^{(t)}_{ki} \left( \frac{(\vect{w}^T \vect{z_i} - \mu_k)}{\sigma_k^2} \right) = 0
\end{equation}
Upon simplifying, $\mu_k$ is given by:
\begin{equation}\label{eq31}
\mu_k = \frac{\sum_{i=1}^n \alpha^{(t)}_{ki} (\vect{w}^T \vect{z_i})}{\sum_{i=1}^n \alpha^{(t)}_{ki}}
\end{equation}

\paragraph{Estimating $\vect{\sigma^2}$}
Differentiating the M-step objective w.r.t $\sigma_k^2$ the optimality condition is:
\begin{equation}\label{eq32}
\deriv{\sigma_k^2} Q\left( \vect{\pi}, \vect{\mu}, \vect{\sigma^2}, \vect{w}, \matr{\alpha}^{(t)} \right)  + \deriv{\sigma_k^2} H_2\left( \vect{\pi}, \vect{\sigma^2} \right) = 0
\end{equation}
Now,
\begin{equation}\label{eq33}
\deriv{\sigma_k^2} Q\left( \vect{\pi}, \vect{\mu}, \vect{\sigma^2}, \vect{w}, \matr{\alpha}^{(t)} \right)  = \sum_{i = 1}^n \alpha^{(t)}_{ki} \left\{  \frac{(\vect{w}^T \vect{z_i} - \mu_k)^2}{2 \sigma_k^4} - \frac{1}{2 \sigma_k^2}  \right\}
\end{equation}
and
\begin{equation}\label{eq34}
\deriv{\sigma_k^2} H_2\left( \vect{\pi}, \vect{\sigma^2} \right) = -\frac{(\theta_k + 1)}{\sigma_k^2} + \frac{1}{\gamma_k} \frac{1}{\sigma_k^4}
\end{equation}
Substituting \ref{eq33} and \ref{eq34} in \ref{eq32} we get:
\begin{equation}\label{eq35}
 \sum_{i = 1}^n \alpha^{(t)}_{ki} \left\{  \frac{(\vect{w}^T \vect{z_i} - \mu_k)^2}{2 \sigma_k^4} - \frac{1}{2 \sigma_k^2}  \right\} -\frac{(\theta_k + 1)}{\sigma_k^2} + \frac{1}{\gamma_k} \frac{1}{\sigma_k^4} = 0
\end{equation}
Upon simplification, we get:
\begin{equation}\label{eq36}
\sigma_k^2 = \dfrac{ 2\, {\gamma^{-1}_k} + \sum_{i=1}^n \alpha^{(t)}_{ki} \, (\vect{w}^T \vect{z_i} - \mu_k)^2 }{ 2(\theta_k + 1) + \sum_{i=1}^n \alpha^{(t)}_{ki} }
\end{equation}

\paragraph{Estimating $\vect{w}$}
The Lagrangian function for optimizing $\vect{w}$ is given by:
\begin{equation}\label{eq37a}
\mathcal{L}(\vect{w},\lambda_1,\vect{\lambda_2}) = Q\left( \vect{\pi}, \vect{\mu}, \vect{\sigma^2}, \vect{w}, \matr{\alpha}^{(t)} \right) - \left\{ \lambda_1 (\vect{w}^T \vect{w} - 1) \right\} - \left\{ \vect{\lambda_2}^T \matr{G}^T \vect{w} \right\} 
\end{equation}
Since $\vect{w}$ must satisfy the constraints \ref{eq3}, and since $H_2\left( \vect{\pi}, \vect{\sigma^2} \right)$ does not depend on $\vect{w}$ the optimality condition is given by:
\begin{equation}\label{eq37}
\deriv{\vect{w}} Q\left( \vect{\pi}, \vect{\mu}, \vect{\sigma^2}, \vect{w}, \matr{\alpha}^{(t)} \right) - \deriv{\vect{w}} \left\{ \lambda_1 (\vect{w}^T \vect{w} - 1) \right\} - \deriv{\vect{w}} \left\{ \vect{\lambda_2}^T \matr{G}^T \vect{w} \right\} = \vect{0}
\end{equation}
In the above equation, $\lambda_1$ is the Lagrange multiplier for the constraint $\vect{w}^T \vect{w} = 1$ and $\vect{\lambda_2}$ is a $q \times 1$ Lagrange multiplier vector for the constraint $\matr{G}^T \vect{w} = \vect{0}$.
Substituting the derivatives, we can write the optimality condition as:
\begin{equation}\label{eq38}
\sum_{i = 1}^n \sum_{k = 1}^R \alpha^{(t)}_{ki} \, \left\{ -\frac{(\vect{w}^T \vect{z_i} - \mu_k) \vect{z_i}}{\sigma_k^2} \right\} - 2 \, \lambda_1 \, \vect{w} - \matr{G} \, \vect{\lambda_2} = \vect{0}
\end{equation}
Upon simplification, we can write the optimality condition as:
\begin{equation}\label{eq39}
\vect{b} - \matr{A} \, \vect{w} - 2 \, \lambda_1 \, \vect{w} - \matr{G} \, \vect{\lambda_2} = \vect{0}
\end{equation}
where the vector $\vect{b}$ and matrix $\matr{A}$ are independent of $\vect{w}$ and are given by:
\begin{equation}\label{eq40}
\vect{b} = \sum_{i = 1}^n \sum_{k = 1}^R \frac{ \alpha^{(t)}_{ki} \mu_k }{\sigma_k^2} \, \vect{z_i}
\end{equation}
and
\begin{equation}\label{eq41}
\matr{A} = \sum_{i = 1}^n \sum_{k = 1}^R \frac{\alpha^{(t)}_{ki}}{\sigma_k^2} \, \vect{z_i} \, \vect{z_i}^T
\end{equation}
Premultiplying both sides of \ref{eq39} by $\vect{w}^T$ and noting the constraints on $\vect{w}$ given by \ref{eq3} we get:
\begin{equation}\label{eq42}
\vect{w}^T \vect{b} - \vect{w}^T \matr{A} \vect{w} - 2 \lambda_1 (1) - 0 = 0
\end{equation}
In other words, the $\lambda_1$ is given by:
\begin{equation}\label{eq43}
\lambda_1 = \frac{1}{2} (\vect{w}^T \vect{b} - \vect{w}^T \matr{A} \vect{w})
\end{equation}
Similarly, premultiplying both sides of \ref{eq39} by $\matr{G}^T$ and noting the constraint $\matr{G}^T \vect{w} = \vect{0}$ we get:
\begin{equation}\label{eq44}
\matr{G}^T (\vect{b} - \matr{A} \vect{w}) - 2 \lambda_1 (\vect{0}) - (\matr{G}^T \matr{G}) \vect{\lambda_2} = \vect{0}
\end{equation}
Note that $\matr{G}$ is of size $q \times L$ with $q > L$ and of full column rank $L$. This means that $\matr{G}^T \matr{G}$ is non-singular and invertible. Hence we can solve for $\vect{\lambda_2}$ to get:
\begin{equation}\label{eq45}
\vect{\lambda_2} = (\matr{G}^T \matr{G})^{-1} \, \matr{G}^T \, (\vect{b} - \matr{A} \vect{w})
\end{equation}
Substituting $\lambda_1$ from \ref{eq43} and $\vect{\lambda_2}$ from \ref{eq45} into \ref{eq39} we can re-write the optimality condition as:
\begin{equation}\label{eq46}
\vect{b} - \matr{A} \, \vect{w} - (\vect{w}^T \vect{b} - \vect{w}^T \matr{A} \vect{w}) \, \vect{w} - \matr{G} \, (\matr{G}^T \matr{G})^{-1} \, \matr{G}^T \, (\vect{b} - \matr{A} \vect{w}) = \vect{0}
\end{equation}
Let
\begin{equation}\label{eq47}
\matr{P_G} = \matr{I_q} - \matr{G} \, (\matr{G}^T \matr{G})^{-1} \, \matr{G}^T
\end{equation}
Essentially $\matr{P_G}$ is an orthogonal projector to the columns of $\matr{G}$. Thus the optimality condition for $\vect{w}$ can be simplified to:
\begin{equation}\label{eq48}
\matr{P_G} \, (\vect{b} - \matr{A} \vect{w}) - (\vect{w}^T \vect{b} - \vect{w}^T \matr{A} \vect{w}) \, \, \vect{w} = \vect{0}
\end{equation}
Since $\vect{w}$ is a $q \times 1$ vector, for fixed $\vect{b}$ and $\matr{A}$ this is a system of $q$ \textbf{cubic} equations in $q$ variables. 

\subsection{Breaking up the M-step into two parts}

Suppose we have already estimated the current respobsibilities $\matr{\alpha}^{(t)}$ in the E-step. Our goal is to simultaneously solve the M-step equations for $\vect{\pi}, \vect{\mu}, \vect{\sigma^2}$ and $\vect{w}$. Our strategy for solving the M-step equations is to note that with $\vect{w}$ fixed, $\vect{\pi}, \vect{\mu}$ and $\vect{\sigma^2}$ can be solved explicitly using \ref{eq28}, \ref{eq31} and \ref{eq36} respectively. While for fixed $\vect{\pi}, \vect{\mu}$ and $\vect{\sigma^2}$ (i.e., fixed $\vect{b}$ and $\matr{A}$), we can calculate $\vect{w}$ by solving a system of cubic equations \ref{eq48}. Hence we can break up the M-step into 2 parts as follows:
\begin{enumerate}
\item \textbf{M-step, Part 1:}\\ In the first part, we keep $\vect{w}$ fixed at its current value $\vect{w}^{*}$ and solve the following maximization problem:
\begin{equation}\label{mstep_part1}
\vect{\pi}^{*}, \vect{\mu}^{*}, \vect{\sigma^2}^{*} = \mbox{arg} \max_{\vect{\pi}, \vect{\mu}, \vect{\sigma^2}} Q\left( \vect{\pi}, \vect{\mu}, \vect{\sigma^2}, \vect{w}^{*}, \matr{\alpha}^{(t)} \right) + H_2\left( \vect{\pi}, \vect{\sigma^2} \right)
\end{equation}
This is equivalent to solving for $\vect{\pi}, \vect{\mu}$ and $\vect{\sigma^2}$ using \ref{eq28}, \ref{eq31} and \ref{eq36}.  

\item \textbf{M-step, Part 2:}\\ In the second part, we keep $\vect{\pi}, \vect{\mu}$ and $\vect{\sigma^2}$ fixed at their current values $\vect{\pi}^{*},\vect{\mu}^{*},\vect{\sigma^2}^{*}$ and 
solve the maximization problem:
\begin{equation}\label{mstep_part2}
\vect{w}^{*} = \mbox{arg} \max_{\vect{w}} \, Q\left( \vect{\pi}^{*}, \vect{\mu}^{*}, \vect{\sigma^2}^{*}, \vect{w}, \matr{\alpha}^{(t)} \right) + H_2\left( \vect{\pi}^{*}, \vect{\sigma^2}^{*} \right)
\end{equation}
This is equivalent to solving the system of cubic equations \ref{eq48} for $\vect{w}$.

\end{enumerate}
We repeatedly perform the alternating maximizations in part 1 and part 2 until the absolute value of the change in $H(\vect{\pi}, \vect{\mu}, \vect{\sigma^2}, \vect{w})$ from one alternating maximization to the next is below a user specified tolerance $\varepsilon_M$. We used $\varepsilon_M = 10^{-3}$ in out experiments.

There is a however a complication that needs to be taken care of when implementing the overall EM step. Note that when excluding $\vect{w}$, the M-step objective has closed form solutions for $\vect{\pi}, \vect{\mu}$ and $\vect{\sigma^2}$. In addition, we can show that these closed form solutions result in a maximization of the partial M-step function with fixed $\vect{w}$. Can we say the same thing about maximizing $\vect{w}$ for fixed $\vect{\pi}, \vect{\mu}$ and $\vect{\sigma^2}$? Does the 2nd part of M-step for optimizing $\vect{w}$ converge to a maximum? Suppose $\vect{w}^*$ is a solution to \ref{eq48}. The Hessian of the Lagrangian \ref{eq37a} is given by:
\begin{equation}\label{eq49}
\nabla^2_{\vect{w} \vect{w}} \, \mathcal{L}( \vect{w}^*, \lambda_1^*, \vect{\lambda_2}^* ) = -\matr{A} - 2 \, \lambda_1^* \matr{I_q}
\end{equation}
Here, $\lambda_1^*$ is simply the Lagrange multiplier from \ref{eq43} evaluated at $\vect{w}^*$. Note that $\mathbf{A}$ is a symmetric and positive definite matrix. If $\lambda_1^* > 0$ then we can guarantee that $\nabla^2_{\vect{w} \vect{w}} \, \mathcal{L}( \vect{w}^*, \lambda_1^*, \vect{\lambda_2}^* )$ will be negative definite and the 2nd part of M-step will have found a local maximum. How can we be sure that $\lambda_1^* > 0$? How can we ensure that the M-step solution for $\vect{\pi}, \vect{\mu}, \vect{\sigma^2}$ and $\vect{w}$ is admissible?

One way to solve this problem is to check \ref{eq23a} explicitly after convergence of each overall M-step. If \ref{eq23a} is not satisfied then we re-initialize the M-step with a random vector $\vect{w}$ that satisfies $\matr{G}^T \vect{w} = 0$ and $||\vect{w}||_2 = 1$ and re-solve the 2 part M-step. This implies we are simply checking for an increase in the objective function \ref{eq23} after each M-step.

\subsection{Solving for the projection vector }

Solving for $\vect{w}$ is equivalent to finding zeros of the set of equations:
\begin{equation}\label{eq50}
\vect{f}(\vect{w}) = \matr{P_G} \, (\vect{b} - \matr{A} \vect{w}) - (\vect{w}^T \vect{b} - \vect{w}^T \matr{A} \vect{w}) \, \, \vect{w} 
\end{equation}
that satisfy the constraints in \ref{eq3}. We consider below three possible cases:
\subsubsection{Case 1}
Suppose $(\vect{b} - \matr{A} \vect{w}) = \vect{0}$. Note that if $\vect{w}$ satisfies $(\vect{b} - \matr{A} \vect{w}) = \vect{0}$ then equation $\vect{f}(\vect{w}) = \vect{0}$ is satisfied. Since $\matr{A}$ is invertible, define a candidate solution for the second part of the M-step as:
\begin{equation}\label{eq50a}
\vect{w_1} = \matr{A}^{-1} \, \vect{b}
\end{equation}
Is this an acceptable solution? If $\vect{w_1}$ satisfies $\vect{w_1}^T \vect{w_1} = 1$ and $\matr{G}^T \, \vect{w_1} = \vect{0}$ then it is an acceptable solution. However, there is no guarantee that this will be true for $\vect{w_1}$ satisfying \ref{eq50a}. Thus the solution \ref{eq50a} for $\vect{w_1}$ is not admissible.

\subsubsection{Case 2}
Suppose $(\vect{w}^T \vect{b} - \vect{w}^T \matr{A} \vect{w}) = \vect{0}$. If $(\vect{w}^T \vect{b} - \vect{w}^T \matr{A} \vect{w}) = \vect{0}$ then solving $\vect{f}(\vect{w}) = \vect{0}$ reduces to solving:
\begin{equation}\label{eq50b}
\matr{P_G} \, (\vect{b} - \matr{A} \vect{w}) = \vect{0}
\end{equation}
Another candidate solution for the second part of the M-step is given by:
\begin{equation}\label{eq50c}
\vect{w_2} = \matr{A}^{-1} \, (\vect{b} - \matr{G} \vect{\eta} )
\end{equation}
where $\vect{\eta}$ is an arbitrary $L \times 1$ vector. Does this $\vect{w_2}$ satisfy the $(L + 1)$ distinct constraints $\vect{w_2}^T \vect{w_2} = 1$ and $\matr{G}^T \vect{w_2} = \vect{0}$? The number of variables on the right hand side of \ref{eq50c} is $L$ but the number of constraints that they have to satisfy is $(L+1)$ (constraints on $\vect{w_2}$) and so the solution \ref{eq50c} for $\vect{w_2}$ is not admissible.

\subsubsection{Case 3}
Suppose $(\vect{w}^T \vect{b} - \vect{w}^T \matr{A} \vect{w}) \neq \vect{0}$ and we solve for $\vect{w}$ that satisfies $\vect{f}(\vect{w}) = 0$. This means that $\vect{w}$ will satisfy:
\begin{equation}\label{eq50d}
\matr{P_G} \, (\vect{b} - \matr{A} \vect{w}) = (\vect{w}^T \vect{b} - \vect{w}^T \matr{A} \vect{w}) \, \, \vect{w} 
\end{equation}
Will such a $\vect{w}$ satisfy $\vect{w}^T \vect{w} = 1$ and $\matr{G}^T \vect{w} = 0$? Premultiplying both sides of \ref{eq50d} by $\matr{G}^T$ and noting the definition of $\matr{P_G}$ in \ref{eq47} we get:
\begin{equation}\label{eq50e}
\matr{G}^T \matr{P_G} \, (\vect{b} - \matr{A} \vect{w}) = \vect{0} = (\vect{w}^T \vect{b} - \vect{w}^T \matr{A} \vect{w}) \, \, \matr{G}^T \vect{w} 
\end{equation}
Since $(\vect{w}^T \vect{b} - \vect{w}^T \matr{A} \vect{w}) \neq \vect{0}$, \ref{eq50e} gives us:
\begin{equation}\label{eq50f}
\matr{G}^T \vect{w} = \vect{0}
\end{equation}
Premultiplying both sides of \ref{eq50d} by $\vect{w}^T$ we get:
\begin{equation}\label{eq50g}
\vect{w}^T \matr{P_G} \, (\vect{b} - \matr{A} \vect{w}) = (\vect{w}^T \vect{b} - \vect{w}^T \matr{A} \vect{w}) \, \, \vect{w}^T \vect{w} 
\end{equation}
From \ref{eq47} and \ref{eq50f} we know that $\vect{w}^T \matr{P_G} = \vect{w}^T$ and so \ref{eq50g} can be written as:
\begin{equation}\label{eq50h}
\vect{w}^T \, (\vect{b} - \matr{A} \vect{w}) = (\vect{w}^T \vect{b} - \vect{w}^T \matr{A} \vect{w}) = (\vect{w}^T \vect{b} - \vect{w}^T \matr{A} \vect{w}) \, \, \vect{w}^T \vect{w} 
\end{equation}
Since $(\vect{w}^T \vect{b} - \vect{w}^T \matr{A} \vect{w}) \neq \vect{0}$, \ref{eq50h} gives us:
\begin{equation}\label{eq50i}
\vect{w}^T \vect{w} = 1
\end{equation}
Therefore if $(\vect{w}^T \vect{b} - \vect{w}^T \matr{A} \vect{w}) \neq \vect{0}$ then any solution to $\vect{f}(\vect{w}) = \vect{0}$ will satisfy both $\vect{w}^T \vect{w} = 1$ and $\matr{G}^T \vect{w} = 0$.
To solve for $\vect{w}$ such that $\vect{f}(\vect{w}) = \vect{0}$ we can use Newton's method. The Jacobian of $\vect{f}$ w.r.t $\vect{w}$ is given by:
\begin{equation}\label{eq51}
\matr{J}(\vect{w}) = -\matr{P_G} \, \matr{A} - \vect{w} \, \vect{b}^T - (\vect{b}^T \vect{w}) \, \matr{I_q} + (\vect{w}^T \matr{A} \vect{w}) \, \matr{I_q} + 2 \, \vect{w}\vect{w}^T \, \matr{A}
\end{equation}
Assuming, $\matr{J}(\vect{w})$ is non-singular the basic Newton update is given by:
\begin{equation}\label{eq52}
\vect{w}^{(i+1)} = \vect{w}^{(i)} - \left[ \matr{J}\left( \vect{w}^{(i)} \right) \right]^{-1} \, \vect{f} \left( \vect{w}^{(i)} \right) 
\end{equation}
To avoid problems caused by intermediate singularity of $\matr{J}$ we can replace $\left[ \matr{J}\left( \vect{w}^{(i)} \right) \right]^{-1}$ by a modified version $\left[ \matr{J}\left( \vect{w}^{(i)} \right) + \eta \matr{I_q} \right]^{-1}$ for a sufficiently large $\eta$ such that $\left[ \matr{J}\left( \vect{w}^{(i)} \right) + \eta \matr{I_q} \right]$ is non-singular (see \cite{Nocedal:book} for more details). For instance, we could use the Levenberg-Marquardt modification of the Newton's method which essentially does this replacement. In this case, the update equation for $\vect{w}$ can be written as:
\begin{equation}\label{eq53}
\vect{w}^{(i+1)} = \vect{w}^{(i)} - \left[ \matr{J}\left( \vect{w}^{(i)} \right) + \eta \matr{I_q} \right]^{-1} \, \vect{f} \left( \vect{w}^{(i)} \right) 
\end{equation}
The projection vector $\vect{w}$ is initialized for the Newton iteration as follows:
\begin{align}\label{newton_init}
\vect{w}^{init} = \matr{P_G} \, \matr{A}^{-1} \, \vect{b} \\
\vect{w}^{init} = \frac{ \vect{w}^{init} }{ ||\vect{w}^{init}||_2 }
\end{align}
The complete EM algorithm pseudocode is given in Fig. \ref{alg1}. 

\begin{figure}
\rule{\textwidth}{1pt}
\textbf{EM algorithm for estimating the PMOG model}
\begin{algorithmic}[1]
\REQUIRE $q \times n$ matrix $\matr{Z}$, number of Gaussian components in MOG $R$, $q \times L$ matrix $\matr{G}$ ($L < q$), Dirichlet prior parameter vector $\vect{\beta}$ for $\vect{\pi}$, Inverse Gamma prior parameter vectors $\vect{\theta}$ and $\vect{\gamma}$ for $\vect{\sigma^2}$ and convergence tolerances $\varepsilon_{rel}$, $\varepsilon_M$
\STATE Select  $\vect{w}^{(0)}$ randomly such that $\matr{G}^T \vect{w}^{(0)} = \vect{0}$ and $|| \vect{w}^{(0)}||_2 = 1$. Initialize the $R \times 1$ vectors $\vect{\pi}^{(0)}, \vect{\mu}^{(0)}, \vect{\sigma^2}^{(0)}$
using the $k$-means algorithm on the projected points $\matr{Z}^T \vect{w}^{(0)}$ and set $found = 0$ and $t = 0$
\WHILE{ $found = 0$ }
\STATE \textbf{E-step}:  Calculate the responsibilities $\matr{\alpha}^{(t)}$ using the current parameter estimates $\vect{\pi}^{(t)}, \vect{\mu}^{(t)}, \vect{\sigma^2}^{(t)}, \vect{w}^{(t)}$ and equation \ref{eq21}.
\STATE To start the \textbf{M-step}, set $\vect{w}^* = \vect{w}^{(t)}$. 
\STATE \textbf{M-step, Part 1}: Given $\matr{\alpha}^{(t)}$ and $\vect{w}^*$ optimize the M-step objective function w.r.t $\vect{\pi}, \vect{\mu}$ and $\vect{\sigma^2}$:
\begin{equation}\label{algEM_eq1}
\vect{\pi}^{*}, \vect{\mu}^{*}, \vect{\sigma^2}^{*} = \mbox{arg} \max_{\vect{\pi}, \vect{\mu}, \vect{\sigma^2}} Q\left( \vect{\pi}, \vect{\mu}, \vect{\sigma^2}, \vect{w}^{*}, \matr{\alpha}^{(t)} \right) + H_2\left( \vect{\pi}, \vect{\sigma^2} \right)
\end{equation}
This problem has an explicit solution given by equations \ref{eq28}, \ref{eq31} and \ref{eq36}.
\STATE \textbf{M-step, Part 2}: Given $\matr{\alpha}^{(t)}$ and $\vect{\pi}^{*},\vect{\mu}^{*},\vect{\sigma^2}^{*}$ optimize the M-step objective function w.r.t $\vect{w}$:
\begin{equation}\label{algEM_eq2}
\vect{w}^{*} = \mbox{arg} \max_{\vect{w}} \, Q\left( \vect{\pi}^{*}, \vect{\mu}^{*}, \vect{\sigma^2}^{*}, \vect{w}, \matr{\alpha}^{(t)} \right) + H_2\left( \vect{\pi}^{*}, \vect{\sigma^2}^{*} \right)
\end{equation}
Solving \ref{algEM_eq2} is equivalent to finding the roots of the equation \ref{eq50}
%\begin{equation}\label{algEM_eq3}
%\vect{f}(\vect{w}) = \matr{P_G} \, (\vect{b} - \matr{A} \vect{w}) - (\vect{w}^T \vect{b} - \vect{w}^T \matr{A} \vect{w}) \, \, \vect{w} = \vect{0}
%\end{equation}
which can be done with Newton or quasi-Newton techniques as in \ref{eq52} and \ref{eq53}.  We initialize $\vect{w}$ as per \ref{newton_init} and solve \ref{eq50} to give $\vect{w}^{*}$.
Steps 1 and 2 are alternated repeatedly in each cycle until the absolute value of the change in $H(\vect{\pi}^{*}, \vect{\mu}^{*}, \vect{\sigma^2}^{*}, \vect{w}^{*})$ from one cycle to the next is $< \varepsilon_M$. Since \ref{algEM_eq2} is a non-concave maximization problem, the solution to \ref{eq50} might converge to a minimum instead of a maximum. Hence we iterate as follows:
	
	\STATE $validsolution = 0$
	\WHILE{ $validsolution = 0$ }
	\IF{equation \ref{eq23a} is satisfied}
		\STATE $validsolution = 1$, set $\vect{\pi}^{(t+1)} = \vect{\pi}^{*}$, $\vect{\mu}^{(t+1)} = \vect{\mu}^{*}$, $\vect{\sigma^2}^{(t+1)} = \vect{\sigma^2}^{*}$, $\vect{w}^{(t+1)} = \vect{w}^{*}$
	\ELSE
	         \STATE Initialize $\vect{w}^{*}$ randomly such that $\matr{G}^T \vect{w}^{*} = \vect{0}$ and $|| \vect{w}^{*} ||_2 = 1$. With this initialization re-solve the 2-part \textbf{M-step}
	\ENDIF
	\ENDWHILE
\STATE If $\mbox{abs} \left[ H(\vect{\pi}^{(t+1)}, \vect{\mu}^{(t+1)}, \vect{\sigma^2}^{(t+1)}, \vect{w}^{(t+1)}) - H( \vect{\pi}^{(t)}, \vect{\mu}^{(t)}, \vect{\sigma^2}^{(t)}, \vect{w}^{(t)}) \right] \le \varepsilon^{*}$, set $found = 1$. In the above equation $\varepsilon^{*} = \varepsilon_{rel} \, \left( \underset{t}{\mbox{mean}} \,\, \mbox{abs} \left[ H( \vect{\pi}^{(t)}, \vect{\mu}^{(t)}, \vect{\sigma^2}^{(t)}, \vect{w}^{(t)}) \right] \right)$.
\STATE $t \gets t + 1$
\ENDWHILE
\end{algorithmic}
%\rule{\textwidth}{1pt}
\caption{EM algorithm for estimating the PMOG model.}
\label{alg1}
\end{figure}

\section{The BSS problem}\label{bss_problem}
In this section, we will try to answer the following two questions;
\begin{itemize}
\item What is the BSS problem and why is its solution difficult?
\item What is the connection between BSS, differential entropy and source correlation?
\end{itemize}

\subsection{The linear BSS problem}
We consider the general version of a linear BSS problem. Vectors $\vect{x_i} \in \mathbf{R}^p$ are generated from latent source vectors $\vect{s_i} \in \mathbf{R}^q$ as follows: 
\begin{equation}\label{bss1}
\vect{x_i} = \vect{\mu} + \matr{A} \, \vect{s_i} + \vect{\varepsilon_i}
\end{equation}
In this linear mixing model, $\matr{A}$ is the $p \times q$ mixing matrix with $p > q$, $\vect{\mu}$ is the $p \times 1$ mean vector and $\vect{\varepsilon_i}$ is the $p \times 1$ noise vector with distribution $\svn{\vect{\varepsilon_i}}{\vect{0}}{\sigma^2 \matr{I}_p}$. We assume without loss of generality that each component of $\vect{s_i}$ has zero mean and unit variance i.e., $E(s_{ij}) = 0$ and $E(s_{ij}^2) = 1$. Thus, the second order statistics of $\vect{s_i}$ can be summarized as:
\begin{gather}\label{bss2}
E(\vect{s_i}) = 0 \\
E(\vect{s_i} \vect{s_i}^T) = \matr{\Sigma_s}
\end{gather}
where $\matr{\Sigma_s}$ is the unknown $q \times q$ correlation matrix between the source components. Note that the diagonal elements of $\matr{\Sigma_s}$ are ones (1's). No distributional assumptions are made on the density of vector $\vect{s_i}$. One simply requires that the components of vector $\vect{s_i}$ are minimally dependent (or maximally independent) on each other in an information theoretic sense. Given $n$ independent realizations of the vector $\vect{x_i}$ (generated from $n$ independent realizations of sources $\vect{s_i}$), the goal is to estimate the unknown sources $\vect{s_i}$ when $\matr{A}, \vect{\mu}$ and the noise variance $\sigma^2$ are unknown. 

\subsection{Why is the BSS problem difficult?}
The BSS problem is difficult because the estimation of mixing parameters $\matr{A}, \vect{\mu}$ and $\sigma^2$ is coupled with the estimation of latent sources $\vect{s_i}$. Note that the equation \ref{bss1} gives us the conditional density of $\vect{x_i}$ given $\vect{s_i}$ as:
\begin{equation}
P( \vect{x_i} \mid \vect{s_i} ) \sim \mathcal{N}\left( \vect{x_i} \mid \vect{\mu} + \matr{A} \, \vect{s_i}, \sigma^2 \matr{I}_p \right)
\end{equation}

The standard approach of estimating mixing parameters by maximum likelihood (ML) would require the computation of marginal density $P(\vect{x_i})$. If the joint source density is  $P(\vect{s_i})$, then we can write:
\begin{equation}\label{complicated1}
P(\vect{x_i}) = \int P(\vect{x_i} \mid \vect{s_i}) \, \, P(\vect{s_i}) \, \vect{ds_i}
\end{equation}
If the above integral is tractable then we can compute $P(\vect{x_i})$ and susequently the ML solution for $\vect{\mu}, \matr{A}$ and $\sigma^2$. However, we are not given any parametric form for $P(\vect{s_i})$. We are simply given that the components of $\vect{s_i}$ are maximally independent with 0 mean, unit-variance and unknown correlation structure $\matr{\Sigma}_s$. This loose specification of $P(\vect{s_i})$ is the root cause of difficulty in the BSS problem. Even if the mixing parameters $\vect{\mu}, \matr{A}$ and $\sigma^2$ are known the computation of posterior mean or maximum aposteriori (MAP) estimate of $\vect{s_i}$ would require some specification of the density of $\vect{s_i}$ since
\begin{equation}\label{complicated2}
P( \vect{s_i} \mid \vect{x_i} ) \propto P( \vect{x_i} \mid \vect{s_i} ) \,\, P( \vect{s_i} )
\end{equation}
Thus it is clear that the BSS problem is non-trivial.

\subsection{Measuring dependence in BSS}
We start this subsection with some fundamental definitions from information theory followed by a detailed study of mutual information as a contrast function for BSS.\\

\begin{defn}
\textbf{Kullback-Leibler divergence:}
Given probability density functions $P(\vect{y})$ and $Q(\vect{y})$, the Kullback-Leibler divergence (or KL distance) between $P$ and $Q$ is given by:
\begin{equation}\label{dependence1}
KL(P, Q) = \int P(\vect{y}) \log \left[ \frac{P(\vect{y})}{ Q(\vect{y}) }\right] \, \vect{dy}
\end{equation}
For any $P$ and $Q$ it is true that $KL(P,Q) \ge 0$.\\
\end{defn}

\begin{defn}
\textbf{Differential entropy:}
Given a random variable $\vect{y}$ with density $P(\vect{y})$, the differential entropy of $\vect{y}$ is defined to be:
\begin{equation}\label{dependence1a}
H(\vect{y}) = - \int P(\vect{y}) \log{P(\vect{y})} \, \vect{dy}
\end{equation}
\end{defn}

\begin{defn}
\textbf{Non-Gaussianity:}
A concept related to differential entropy is the non-Gaussianity (NG) of a distribution. Given a $q \times 1$ random vector $\vect{y}$ with density $P(\vect{y})$, mean $\vect{\mu_y}$ and co-variance $\matr{\Sigma_y}$, suppose $\mathcal{N}(\vect{y} \mid \vect{\mu_y}, \matr{\Sigma_y})$ is a Normal density with the same mean and co-variance as $\vect{y}$. Then the non-Gaussianity of $\vect{y}$ is defined to be (see \cite{Cardoso:2003}):
\begin{equation}\label{dependence1b}
NG(\vect{y}) = KL\left( P(\vect{y}), \mathcal{N}(\vect{y} \mid \vect{\mu_y}, \matr{\Sigma_y}) \right)
\end{equation}
An important property of non-Gaussianity is invariance to invertible linear transformations i.e., 
\begin{equation}\label{dependence1b1}
NG(\vect{a} + \matr{B} \vect{y}) = NG(\vect{y}) 
\end{equation}
for any non-singular matrix $\matr{B}$ and vector $\vect{a}$.\\
\end{defn}

\begin{rem}
\textbf{Differential entropy and non-Gaussianity:}
Invoking the definition of KL divergence in \ref{dependence1b}, we get:
\begin{equation}\label{dependence1c}
NG(\vect{y}) =  \int P(\vect{y}) \log \left[ \frac{P(\vect{y})}{ \mathcal{N}(\vect{y} \mid \vect{\mu_y}, \matr{\Sigma_y} ) }\right] \, \vect{dy}
\end{equation}
After some algebraic manipulations and noting that $\vect{y}$ has co-variance $\matr{\Sigma_y}$ we get:
\begin{equation}\label{dependence1d}
NG(\vect{y}) = - H(\vect{y}) + \frac{q}{2} \log{2 \pi e} + \frac{1}{2} \log [ \det (\matr{\Sigma_y}) ] 
\end{equation}
Note that in equation \ref{dependence1d}, $q$ is the length of the vector $\vect{y}$.\\
\end{rem}

\begin{defn}
\textbf{Mutual information:}
An information theoretic measure of the dependence between components of a $q \times 1$ random vector $\vect{s}$ with density $P(\vect{s})$ is the KL distance between $P(\vect{s})$ and the density of $\vect{s}$ when its components are independent i.e., $P^{id}(\vect{s}) = \prod_{j = 1}^q P_j( s_j )$ where $P_j$ is the marginal density of $s_j$. This KL distance is also called the mutual information \cite{Cardoso:2003} between the components of $\vect{s}$:
\begin{equation}\label{dependence2}
I(\vect{s}) = KL( P(\vect{s}), P^{id}(\vect{s}) )
\end{equation}
Substituting the definitions of KL distance and $p^{id}$ and simplifying we get the classical expression for mutual information between components of $\vect{s}$:
\begin{equation}\label{dependence3}
I(\vect{s}) = \sum_{j = 1}^q H(s_j) - H(\vect{s})
\end{equation}
\end{defn}

Suppose the vector $\vect{s}$ is generated from another $q \times 1$ vector $\vect{z}$ with density $P(\vect{z})$ via a non-singular linear transformation:
\begin{equation}\label{dependence4}
\vect{s} = \matr{W}^T \vect{z} = \begin{pmatrix} \vect{w_1}^T \\ \vect{w_2}^T \\ \vdots \\ \vect{w_q}^T \end{pmatrix} \vect{z}
\end{equation}
then we can re-write equation \ref{dependence3} as:
\begin{equation}\label{dependence5}
I(\vect{s}) = \sum_{j = 1}^q H(s_j) - H(\vect{z}) - \log [ \mbox{abs}( \det (\matr{W}^T) ) ]
\end{equation}
where $H(\vect{z})$ is the differential entropy of $\vect{z}$. If we are given that $\vect{s}$ satisfies $E(\vect{s}) = \vect{0}$ and $E(\vect{s} \vect{s}^T) = \matr{\Sigma_s}$ with $1$'s on the diagonal then $\vect{z}$ must have $\vect{0}$ mean and co-variance $\matr{\Sigma_z}$ satisfying:
\begin{equation}\label{dependence6}
\matr{\Sigma_s} = \matr{W}^T \, \matr{\Sigma_z} \, \matr{W}
\end{equation}
Taking determinants on both sides, it is easy to see that:
\begin{equation}\label{dependence7}
\mbox{abs}( \det(\matr{W}^T) ) = \frac{\det(\matr{\Sigma_s})^{\frac{1}{2}}}{\det(\matr{\Sigma_z})^{\frac{1}{2}}}
\end{equation} 
Substituting \ref{dependence7} into \ref{dependence5} we see that:
\begin{equation}\label{dependence8}
I(\vect{s}) = \underbrace{ \left\{ \sum_{j = 1}^q H(s_j) - \frac{1}{2} \log [ \det (\matr{\Sigma_s}) ] \right\} }_{\mbox{dependent on $\matr{W}^T$}}  + \underbrace{ \left\{ \frac{1}{2} \log [ \det (\matr{\Sigma_z}) ] - H(\vect{z})  \right\} }_{\mbox{dependent only on $P(\vect{z})$}}
\end{equation}
This equation is identical to equation (16) in \cite{Cardoso:2003}. This can be seen by replacing differential entropy by non-Gaussianity in \ref{dependence8} using \ref{dependence1d}, noting that individual components $s_j$ of $\vect{s}$ have unit variance and using the invariance property \ref{dependence1b1} for non-Gaussianity. Note that the diagonal elements of $\matr{\Sigma_s}$ are ones and so the correlation $C(\vect{y})$ as defined in \cite{Cardoso:2003} is equivalent to $- \frac{1}{2} \log [ \det (\matr{\Sigma_s}) ]$. It is clear that given a density $P(\vect{z})$ for $\vect{z}$, the second term is independent of the linear transformation $\matr{W}^T$. Thus the dependence between components of $\vect{s}$ is fully captured by the first term alone which depends on $\vect{W}^T$ (see \ref{dependence6} and \ref{dependence4}).\\

\begin{rem}
\textbf{Note on the $- \frac{1}{2} \log [ \det (\matr{\Sigma_s}) ] $ term}:
First, note that $\matr{\Sigma_s}$ is a correlation matrix with 1s on the diagonal. If $\phi_1, \phi_2,\ldots,\phi_q$ are its eigenvalues then it follows that
\begin{gather}\label{note1}
\mbox{trace}(\matr{\Sigma_s}) = q = \sum_{i = 1}^q \phi_i \\
\det (\matr{\Sigma_s}) = \prod_{i = 1}^q \phi_i
\end{gather}
Therefore it follows that:
\begin{align}\label{note2}
\log [ \det (\matr{\Sigma_s}) ] & = \sum_{i = 1}^q \log \phi_i \\
& = q \left[ \frac{1}{q} \sum_{i = 1}^q \log \phi_i \right] \\
& \le q \log \left[ \frac{1}{q} \sum_{i = 1}^q \phi_i \right] && \text{by concavity of $\log$ function} \\
& = q \log \left[ \frac{1}{q} q \right] && \text{using the trace condition in \ref{note1}} \\
& = 0
\end{align}
Thus we see that
\begin{align}\label{note3}
\log [ \det (\matr{\Sigma_s}) ] &\le 0 \, \mbox{ or } \\
- \frac{1}{2} \log [ \det (\matr{\Sigma_s}) ]  &\ge 0
\end{align}
The minimal value of $- \frac{1}{2} \log [ \det (\matr{\Sigma_s}) ]$ is 0 and is attained when $\det (\matr{\Sigma_s}) = 1$. Since $\matr{\Sigma_s}$ is a correlation matrix with 1s on the diagonal the only way $\det (\matr{\Sigma_s}) = 1$ can hold is if $\matr{\Sigma_s} = \matr{I}_q$. \begin{center}\fbox{The term $- \frac{1}{2} \log [ \det (\matr{\Sigma_s}) ] $ attains a minimal value of 0 when $\matr{\Sigma_s} = \matr{I}_q$.}\end{center}
\end{rem}

The optimization problem for minimizing dependence between components of $\vect{s}$ can therefore be written as:
\begin{align}\label{dependence9}
\mbox{min}_{\matr{W}^T} \,\,\,\,  f(\vect{s}) &=  \underbrace{ \left\{ \sum_{j = 1}^q H(s_j) \right\} }_{ \mbox{encourages non-Gaussianity} } + \,\,\,\, \psi \, \underbrace{ \left\{- \frac{1}{2} \log [ \det (\matr{\Sigma_s}) ] \right\} }_{ \mbox{$\ge 0$, encourages $\matr{\Sigma}_s = \matr{I}_q$} } \\
\mbox{ where: } & \\
 s_j &= \vect{w_j}^T \vect{z}, \,\,\ \mbox{abs}( \det(\matr{W}^T) ) = \frac{\det(\matr{\Sigma_s})^{\frac{1}{2}}}{\det(\matr{\Sigma_z})^{\frac{1}{2}}} \\
 \mbox{ subject to: } & \\
\matr{\Sigma_s} & = \matr{W}^T \, \matr{\Sigma_z} \, \matr{W}
\end{align}
Since the sources $s_j$ have unit variance, using the property \ref{dependence1d} we get:
\begin{equation}\label{dependence10}
NG(s_j) = - H(s_j) + \frac{1}{2} \log{2 \pi e}
\end{equation}
Minimizing $f(\vect{s})$ is therefore a problem of minimizing the sum of 2 different terms. The first term measures the differential entropy of each component $s_j$. As seen from \ref{dependence10}, minimizing the differential entropy under the unit variance constraint is equivalent to maximizing the non-Gaussianity $NG(s_j)$. Thus minimizing the first term encourages non-Gaussianity. The second term measures the 2nd order cross-correlation between sources. Since this term is minimized when sources are uncorrelated, it encourages uncorrelatedness of the sources. The weighting constant $\psi$ serves to balance the two terms in the objective function. It is instructive to note the form of the objective function for various values of $\psi$:
\begin{itemize}
\item When $\psi = 0$, the 2nd term drops out. In this case, minimization of differential entropy (or equivalently maximization of non-Gaussianity) of sources is considered much more important than minimizing correlatedness. Without making any additional assumptions on the correlation structure of $\vect{s}$, the projection vectors $\vect{w_i}$ are constrained by the relation $\vect{w_i}^T \matr{\Sigma_z} \vect{w_i} = 1$.
\item When $\psi = 1$, then minimizing \ref{dependence9} is equivalent to minimization of mutual information $I(\vect{s})$. In this case, equal importance is given to maximizing non-Gaussianity and uncorrelatedness. As before, the projection vectors $\vect{w_i}$ satisfy $\vect{w_i}^T \matr{\Sigma_z} \vect{w_i} = 1$.
\item When $\psi = \infty$, the 2nd term is forced to become 0 at $\matr{\Sigma_s} = \matr{I}_q$. In this case, non-Gaussianity is maximized under the uncorrelated source assumption. This means that the projection vectors are constrained by: $\vect{w_i}^T \matr{\Sigma_z} \vect{w_j} = \delta_{ij}$ where $\delta_{ij} = 1$ if $i = j$ and $0$ otherwise.
\end{itemize}

The objective function for minimization in \ref{dependence9} can also be re-written as:
\begin{align}\label{dependence11}
\mbox{min}_{\matr{W}^T} \,\,\,\,  f(\vect{s}) &=  \underbrace{ \left\{ \sum_{j = 1}^q H(s_j) \right\} }_{\mbox{encourages non-Gaussianity} } + \,\, \psi \, \underbrace{ \left\{- \frac{1}{2} \log [ \det (\matr{\Sigma_z})] - \log [ \mbox{abs} (\det (\matr{W}^T)) ] \right\} }_{ \mbox{$\ge 0$, encourages $\matr{\Sigma}_s = \matr{I}_q$} } \\
\mbox{ where: } & \\
 s_j &= \vect{w_j}^T \vect{z}, \,\,\ \mbox{abs}( \det(\matr{W}^T) ) = \frac{\det(\matr{\Sigma_s})^{\frac{1}{2}}}{\det(\matr{\Sigma_z})^{\frac{1}{2}}} \\
 \mbox{ subject to: } & \\
\matr{\Sigma_s} & = \matr{W}^T \, \matr{\Sigma_z} \, \matr{W}
\end{align}

\section{Solving the BSS problem}\label{bss_approaches}
Solving the BSS problem means estimating both the mixing parameters $\vect{\mu}, \matr{A}$ and $\sigma^2$ and the latent sources $\vect{s_i}$ given $n$ mixed vectors $\vect{x_i}$. 
We will try to answer the following questions in this section:
\begin{itemize}
\item What are the main solution approaches when the sources are assumed to be uncorrelated and how does the work of Hyvarinen et al.\ \cite{negentropy:1998} relate to the problem of BSS?
\item How should the solution approach change when non-zero second order source correlation is allowed?
\end{itemize}

\subsection{Case I: Uncorrelated sources}

In this case, we assume that $\matr{\Sigma_s} = \matr{I}_q$ i.e., the sources are uncorrelated. There are two main solution techniques in this case.

\subsubsection{ (1) The solution by Attias et al.\ under exact independence} 
Attias et al.\ \cite{Attias:1999} in a seminal paper described a general solution to the linear BSS problem under exact independence. He assumed that each component of the vector $\vect{s_i}$ is described by a mixture of Gaussians (MOG) density. The logic for this was that since the MOG is a very flexible density (given sufficient components in the mixture), it should be able to describe more complicated and non-Gaussian source densities. Given the fact that components of $\vect{s_i}$ are exactly independent, we have
\begin{equation}\label{complicated3}
P( \vect{s_i} ) = \prod_{j = 1}^q P_j( s_{ij} )
\end{equation}
where $s_{ij}$ is the $j$th component of vector $\vect{s_i}$ and $P_j$ is the marginal density of the $j$th component of $\vect{s_i}$.  In other words, the source density $P( \vect{s_i} )$ has a very flexible parametric form of a product of $q$ MOG densities. This type of a density for $\vect{s_i}$ is also called a "factorial" MOG which is a special case of a mixture of co-adaptive Gaussians densities. Attias et al.\ also showed that under this parametric form the integral \ref{complicated1} is analytically tractable and the density $P( \vect{x_i} )$ is also a mixture of co-adaptive Gaussians (although not "factorial" MOG as in the case of $\vect{s_i}$). Attias et al.\ also derive an exact EM algorithm for obtaining the ML solution for $\vect{A}, \vect{\mu}, \sigma^2$ as well as the MOG parameters for the source density $P( \vect{s_i} )$. The only limitation of this algorithm is the computational intractability for large problems. Attias et al.\ note that a 13 source mixture with each source described by a 3 component MOG would require $3^{13}$ sums in each E-step, making it computationally intractable. To overcome this problem Attias et al.\ propose a variational approximation to the ML solution instead of the EM algorithm. In summary, an exact solution for the BSS problem under exact independence exists, but is computationally intractable for large problems (greater than ~13 sources) and therefore one has to resort to approximate solutions.

\subsubsection{ (2) PPCA and least squares based solution}
An alternative and much simpler approach is to use an approximation to the density of $\vect{s_i}$ only for the purposes of computing $\matr{A}, \vect{\mu}$ and $\sigma^2$. This approximation should be such that it simplifies computation of the integral \ref{complicated1} making the ML solution possible under the approximate density of $\vect{s_i}$. A density that satisfies this requirement is $P( \vect{s_i} ) = \mvn{ s_i }{ 0 }{ I_q}$. Under this assumption, the model \ref{bss1} simply reduces to the probabilistic PCA (PPCA) model of Tipping et al.\ \cite{Tipping:1999a}. The ML solution for $\matr{A}, \vect{\mu}$ and noise variance $\sigma^2$ under the PPCA model (when $\vect{s_i}$ is Gaussian) is known. Let the $p \times 1$ vector $\bar{\vect{x}}$ be the mean of observations $\vect{x_i}$ and $\matr{S_x}$ be the $p \times p$ sample co-variance i.e., 

\begin{gather}\label{ppca1}
\bar{\vect{x}} = \frac{1}{n} \sum_{i = 1}^n \vect{x_i} \\
\matr{S_x} = \frac{1}{n} \sum_{i = 1}^n (\vect{x_i} - \bar{\vect{x}}) \, (\vect{x_i} - \bar{\vect{x}})^T
\end{gather}

Suppose
\begin{equation}\label{ppca2}
\matr{S_x} = \matr{U} \matr{\Lambda} \matr{U}^T
\end{equation}
is the eigen decomposition of the symmetic matrix $\matr{S_x}$. Here $\matr{U}$ is a $p \times p$ matrix of eigenvectors of $\matr{S_x}$ and $\Lambda$ is a $p \times p$ diagonal matrix containing the corresponding eigenvalues of $\matr{S_x}$.
\begin{equation}\label{ppca3}
\Lambda = \begin{pmatrix} \lambda_1& 0 & \ldots & 0 \\ 0 & \lambda_2 & \ldots & 0 \\ \vdots & \ldots & \ddots & 0 \\ 0 & \ldots & \ldots & \lambda_p \end{pmatrix}
\end{equation}
Suppose we order the eigenvalues of $\matr{S_x}$ such that $\lambda_1 \ge \lambda_2 \ldots \ge \lambda_p \ge 0$. Let $\matr{U_q}$ be a $p \times q$ submatrix of $\matr{U}$ containing the eigenvectors corresponding to the $q$ largest eigenvalues $\lambda_1, \ldots, \lambda_q$ and $\matr{\Lambda_q}$ be the $q \times q$ submatrix of $\Lambda$ with $\lambda_1, \ldots, \lambda_q$ on the diagonal. Then as shown in Tipping et al.\ the ML solution for mixing parameters under the PPCA model is given by:

\begin{gather}\label{ppca3a}
\vect{\hat{\mu}} = \bar{\vect{x}} \\
\hat{\sigma}^2 = \left(\dfrac{1}{p-q}\right)  \sum_{i = q + 1}^p \lambda_i  \\
\matr{\hat{A}} = \matr{U_q} \, (\matr{\Lambda_q} - \hat{\sigma}^2 \matr{I_q})^{1/2} \, \matr{Q}^T \\
\matr{Q}^T \matr{Q} = \matr{I_q}
\end{gather}
The ML solution is unique upto an arbitrary $q \times q$ orthogonal matrix $\matr{Q}$. Following this step, we can estimate the sources $\vect{\hat{s}_i}$ using least squares to get:

\begin{equation}\label{leastsquares1}
\vect{\hat{s}_i} = \matr{Q} \, (\matr{\Lambda_q} - \hat{\sigma}^2 \matr{I_q})^{-\frac{1}{2}} \, \matr{U_q}^T \, (\vect{x_i} - \vect{\hat{\mu}})
\end{equation}

The advantage of this approach is that it decouples the estimation of mixing parameters $\matr{A}, \mu, \sigma^2$ from the sources $\vect{s_i}$. This approach is used for example in Hyvarinen et al.\ (\cite{Hyvarinen:1999}) with $\sigma^2 = 0$ and Beckmann et al.\ (\cite{Beckmann:2004}) with $\sigma^2 \neq 0$.

Now suppose
\begin{gather}\label{bss7}
\matr{z_i} = (\matr{\Lambda_q} - \hat{\sigma}^2 \matr{I_q})^{-\frac{1}{2}} \, \matr{U_q}^T \, (\vect{x_i} - \vect{\hat{\mu}}) \mbox{ and } \\
\matr{Q} = \matr{W}^T = \begin{pmatrix} \vect{w_1}^T \\ \vect{w_2}^T \\ \vdots \\ \vect{w_q}^T \end{pmatrix}
\end{gather}
then
\begin{equation}\label{bss8}
\vect{\hat{s}_i} = \matr{W}^T \vect{z_i} = \begin{pmatrix} \vect{w_1}^T \vect{z_i} \\ \vect{w_2}^T  \vect{z_i} \\ \vdots \\ \vect{w_q}^T \vect{z_i} \end{pmatrix}
\end{equation}

Since $\matr{Q}$ is orthogonal the vectors $\vect{w_i}$ satisfy the constraints
\begin{equation}\label{bss9}
\vect{w_i}^T \vect{w_j} = \delta_{ij}
\end{equation}
where $\delta_{ij} = 1$ if $i = j$ and 0 otherwise. Since $\matr{W}^T$ is orthogonal, the second term in \ref{dependence11} becomes independent of $\matr{W}^T$ and so the objective function $f(\vect{\hat{s}_i})$ reduces to a sum of the differential entropies of the components of $\vect{\hat{s}_i}$:
\begin{align}\label{bss10}
\mbox{min}_{\matr{W}^T} \,\,\,\,  f(\vect{\hat{s}_i}) \,\,\,\, &\propto  \underbrace{ \left\{ \sum_{j = 1}^q H(\hat{s}_{ij}) \right\} }_{ \mbox{encourages non-Gaussianity} } \\
\mbox{ where: } & \\
 \hat{s}_{ij} &= \vect{w_j}^T \vect{z_i}
\end{align}
It is worth noting that since $\matr{W}^T$ is constrained to be orthogonal, no apriori assumptions can be made on the correlation structure $\matr{\Sigma_{\hat{s}_i}}$ of $\vect{\hat{s}_i}$. Thus the components of $\vect{\hat{s}_i}$ will have co-variance $\matr{\Sigma_{\hat{s}_i}} = \matr{W}^T \, \matr{\Sigma_{z_i}} \, \matr{W}$ where $ \matr{\Sigma_{z_i}}$ is the co-variance of $\vect{z_i}$. Now $\matr{\Sigma_{z_i}} = \matr{I}_q$ only if $\hat{\sigma}^2 = 0$. This means that unless $\hat{\sigma}^2 = 0$, the components of $\hat{s}_i$ will \textbf{not} have unit variance and $0$ cross-correlation.

\subsubsection{Differential entropy approximations of Hyvarinen et al.\ }
In a seminal paper \cite{negentropy:1998}, Hyvarinen et al.\ proposed approximations to the differential entropy function $H(x)$ of a random variable with density $P(x)$. The key idea in this work is to approximate $P(x)$ by a maximum entropy distribution (MED) given estimates of the expectations of $m$ functions $G_i$ of $x$ i.e., given $E_x[ G_i(x) ] = c_i$. The solution to this problem is well known \cite{cover:book}:
\begin{equation}\label{maxent1}
p^{med}(x) = a_0 \, \mbox{exp}\left[ \sum_{i = 1}^m a_i \, G_i(x) \right]
\end{equation}
where $a_0, a_1,\ldots, a_m$ are constants. These $(m+1)$ constants can be solved for by simultaneously solving the $m$ expectation equations under the MED density $E_x[ G_i(x) ] = c_i$ together with the normalizing equation $\int p^{med}(x) \, dx = 1$. Simultaneous solution of this  $(m+1)$ system of non-linear equations is difficult. Hence, Hyvarinen et al.\ proposed a solution in which it is assumed that the density $P(x)$ is "not very far from a Gaussian distribution". Under these conditions, using the simplified form of $p^{med}(x)$, Hyvarinen et al.\ derived approximations to the differential entropy $H(x)$ (see \cite{negentropy:1998_techrep} for details). In summary, suppose $x$ has mean $0$ and variance $\sigma^2$. Let $\nu$ be a Gaussian variable with the same mean and variance as $x$. The "near Gaussian" MED density approximation of Hyvarinen et al.\ for a single expectation constraint using a function $\bar{G}(x)$ is given by:
\begin{equation}\label{maxent1a}
p^{med}_{Gaussian}(x) = \mathcal{N}(x \mid 0, \sigma^2) \left\{ 1 + c \, G(x) \right\}
\end{equation}
The function $\bar{G}(x)$ is related to its normalized version $G(x)$ by the equation:
\begin{equation}\label{maxent3}
G(x) = \frac{1}{\delta} ( \bar{G}(x) + \alpha_1 \, x  + \alpha_2 \, x^2 + \gamma )
\end{equation}
Here $\bar{G}(x)$ is any function of $x$ not necessarily even or odd. The 4 constants $\alpha_1, \alpha_2, \gamma$ and $\delta$ are determined from the relations:
\begin{align}\label{maxent4}
\int \mathcal{N}(x \mid 0, \sigma^2) \, x^{k} \, G(x) \,\, dx &= 0, \mbox{ where }  k = 0, 1, 2 \\
\int \mathcal{N}(x \mid 0, \sigma^2) \, G(x) \, G(x) \,\, dx &= 1
\end{align}
Then the differential entropy approximation developed in Hyvarinen et al.\ using $p^{med}_{Gaussian}(x)$ from \ref{maxent1a} is given by:
\begin{equation}\label{maxent2}
H(x) \approx H(\nu) - \frac{1}{2\delta^2} \left\{ E_x[\bar{G}(x)] - E_{\nu}[\bar{G}(\nu)] \right\}^2
\end{equation}
Note that the term $E_{\nu}[\bar{G}(\nu)]$ is $0$ in case $\bar{G}(x)$ is an odd function. Substituting the differential entropy of a Gaussian random variable with mean 0 and variance $\sigma^2$ in \ref{maxent2} we get:
\begin{align}\label{maxent5}
H(x) &\approx \frac{1}{2} \log 2\pi e + \frac{1}{2} \log \sigma^2 - \frac{1}{2\delta^2} \left\{ E_x[\bar{G}(x)] - E_{\nu}[\bar{G}(\nu)] \right\}^2 \\
\nu &= \mbox{ Random variable with density $\mathcal{N}(\nu \mid 0, \sigma^2)$ }\\
\sigma^2 &= \mbox{Var}(x)
\end{align}
Given a set of observed samples of $x$: $x_1,x_2,\ldots,x_n$, the objective function in \ref{maxent5} can be estimated by replacing the expected values by sample averages.

Hyvarinen et al.\ proposed a solution in which the first component of $\vect{\hat{s}_i}$ is estimated by minimizing the differential entropy of the empirical density of $\hat{s}_{i1} = \vect{w_1}^T \vect{z_i}$ over $n$ realizations using the approximation $H(\hat{s}_{i1})$ from \ref{maxent5}.This minimization is carried out under the constraint $\vect{w_1}^T \vect{w_1} = 1$. The second component is estimated similarly by minimizing the differential entropy of the empirical density of $\hat{s}_{i2} = \vect{w_2}^T \vect{z_i}$ over $n$ realizations while imposing the constraint $\vect{w_2}^T \vect{w_1} = 0$ and $\vect{w_2}^T \vect{w_2} = 1$. In general, the $m$th component is estimated by minimizing the differential entropy of the empirical density of $\vect{w_m}^T \vect{z_i}$ over $n$ realizations subject to the constraint:
\begin{gather}\label{bss10a}
\vect{w_m}^T \vect{w_m} = 1 \\
\matr{G}^T \vect{w_m} = 0 \\
\end{gather}
where $\matr{G} = [\vect{w_1},\vect{w_2},\ldots,\vect{w_{m-1}}]$ is a $q \times (m-1)$ matrix with $q > (m-1)$. This is essentially the FastICA (FICA) algorithm proposed by Hyvarinen et al.\ \cite{Hyvarinen:1999, FPICA:1999} and remains the most popular ICA algorithm to date.

\subsection{Case II: Correlated sources}\label{correlated_sources}
Sometimes the assumption of zero second order correlation between the components of $\vect{s_i}$ might be unrealistic. The modified assumption is that of maximal independence between components of $\vect{s_i}$ under unknown and potentially non-zero correlation i.e., $\matr{\Sigma_s} \neq \matr{I}_q$. If we introduce a change of variables: 
\begin{gather}\label{nonorthogonal2}
 \vect{s^*_i}  = \matr{\Sigma_s^{-\frac{1}{2}}} \, \vect{s_i} \\
 \matr{A^*} = \matr{\Sigma_s^{\frac{1}{2}}} \, \matr{A}
\end{gather}
then this case can be reduced to the standard PPCA model:
\begin{align}\label{nonorthogonal2a}
\vect{x_i} & = \vect{\mu} + \matr{A} \, \vect{s_i} + \vect{\varepsilon_i} \\
& =  \vect{\mu} + \matr{A^*} \, \vect{s^*_i} + \vect{\varepsilon_i}
\end{align}
The transformed sources $\vect{s^*_i}$ satisfy $E(\vect{s^*_i}) = 0$ and $E(\vect{s^*_i} \, \vect{s^*_i}^T) = \matr{I_q}$, however maximal independence requirement will be imposed on $\vect{\hat{s}_i}$ and not on $\vect{\hat{s}^*_i}$. Therefore, following the development in the previous section, the PPCA solution is given by:
\begin{gather}\label{nonorthogonal3}
\vect{\hat{\mu}} = \bar{\vect{x}} \\
\hat{\sigma}^2 = \left(\dfrac{1}{p-q}\right)  \sum_{i = q + 1}^p \lambda_i  \\
\matr{\hat{A}^*} = \matr{U_q} \, (\matr{\Lambda_q} - \hat{\sigma}^2 \matr{I_q})^{1/2} \, \matr{Q}^T \\
\matr{Q}^T \matr{Q} = \matr{I_q}
\end{gather}
The least squares source estimates are given by:
\begin{equation}\label{nonorthogonal4}
\vect{\hat{s}^*_i} = \matr{\Sigma_s^{-\frac{1}{2}}} \, \vect{\hat{s}_i}  = \matr{Q} \, (\matr{\Lambda_q} - \hat{\sigma}^2 \matr{I_q})^{-\frac{1}{2}} \, \matr{U_q}^T \, (\vect{x_i} - \vect{\hat{\mu}})
\end{equation}

From \ref{nonorthogonal4} and \ref{bss7}
\begin{align}\label{nonorthogonal4a}
\vect{\hat{s}_i}  &=  \matr{\Sigma_s^{\frac{1}{2}}} \matr{Q} \, (\matr{\Lambda_q} - \hat{\sigma}^2 \matr{I_q})^{-\frac{1}{2}} \, \matr{U_q}^T \, (\vect{x_i} - \vect{\hat{\mu}}) \\
&= \matr{\Sigma_s^{\frac{1}{2}}} \matr{Q} \, \matr{z_i} \\
& = \matr{W^*}^T \, \matr{z_i}
\end{align}
where 
\begin{equation}\label{nonorthogonal5}
\matr{W^*}^T =  \begin{pmatrix} \vect{w^*_1}^T \\ \vect{w^*_2}^T \\ \vdots \\ \vect{w^*_q}^T \end{pmatrix} = \matr{\Sigma_s^{\frac{1}{2}}} \matr{Q}
\end{equation}
is the unmixing matrix for recovering the sources $\vect{\hat{s}_i}$. Note that when the sources are not exactly uncorrelated, the unmixing matrix satisfies $\matr{W^*}^T \matr{W^*} = \matr{\Sigma_s}$ instead of $\matr{W}^T \matr{W} = \matr{I_q}$ as in the exactly uncorrelated case. In other words, sources $\vect{\hat{s}_i}$ are extracted by non-orthogonal projections of vectors $\vect{z_i}$. In this case, the optimization problem can be written as (ignoring terms independent of $\matr{W^*}^T$):

\begin{align}\label{nonorthogonal5a}
\mbox{min}_{\matr{W^*}^T} \,\,\,\,  f(\vect{\hat{s}_i}) &\propto  \underbrace{ \left\{ \sum_{j = 1}^q H(\hat{s}_{ij}) \right\} }_{ \mbox{encourages non-Gaussianity} } + \,\, \psi \, \underbrace{ \left\{- \log [ \mbox{abs} (\det (\matr{W^*}^T)) ] \right\} }_{ \mbox{$\ge 0$, encourages $\matr{\Sigma}_s = \matr{I}_q$} }  \\ 
\mbox{ where: } & \\
 \hat{s}_{ij} &= \vect{w^*_j}^T \vect{z_i}, \,\,\ \mbox{abs}( \det(\matr{W^*}^T) ) = \frac{\det(\matr{\Sigma_{\hat{s}_i}})^{\frac{1}{2}}}{\det(\matr{\Sigma_z})^{\frac{1}{2}}} \\
 \mbox{ subject to: } & \\
\matr{\Sigma_{\hat{s}_i}} & = \matr{W^*}^T \, \matr{\Sigma_z} \, \matr{W^*}
\end{align}

Note that $\matr{\Sigma_{\hat{s}_i}} = \matr{\Sigma_s}$ only if $\matr{\Sigma_z} = \matr{I}_q$. Thus the individual components of $\vect{\hat{s}_i}$ will \textbf{not} have unit variance unless $\hat{\sigma}^2 = 0$. In the case of potentially correlated sources, the 2nd term in the objective function \ref{nonorthogonal5a} does not drop out. Hence sequential extraction of sources is only possible in the case when $\psi = 0$. As discussed before equation \ref{dependence11}, this corresponds to the case when minimization of differential entropy is considered to be much more important than minimizing correlatedness. We consider the solution of \ref{nonorthogonal5a} with $\psi = 0$ in the correlated case.

Since $\matr{W^*}^T \matr{W^*} = \matr{\Sigma_s}$ and the diagonal elements of $\matr{\Sigma_s}$ are ones, the $m$th component is estimated by minimizing the differential entropy of the empirical density of $\hat{s}_{im} = \vect{w^*_m}^T \vect{z_i}$ over $n$ realizations subject to the only constraint:
\begin{gather}\label{nonorthogonal6}
\vect{w^*_m}^T \vect{w^*_m} = 1
\end{gather}
Since the off diagonal elements of $\matr{\Sigma_s}$ are allowed to have non-zero values, no additional mutual orthogonality constraints are imposed on the projection vectors. An important question is "How do you impose non-orthogonality?". It is possible that 2 vectors $\vect{w^*_1}$ and $\vect{w^*_2}$ are identical, since the algorithm does not explicitly enforce non-orthogonality. A simple workaround is to randomly initialize each vector $\vect{w^*_m}$ by making $\vect{w^*_m}$ orthogonal to the previously estimated vectors. In our experiments, this normally prevents convergence to a previously found solution, but again this is not guaranteed. In general, one must continue to randomly initialize solution for the $m$th component until it is found to be different from the previous $(m-1)$ solution vectors.

\section{PMOG based BSS}\label{pmog_based_bss}
In this section, we try to answer the following questions:
\begin{itemize}
\item What is the relationship between the PMOG objective function and differential entropy?
\item How can PMOG be applied to the problem of BSS?
\end{itemize} 

\subsection{PMOG objective and differential entropy}\label{pmog_diffent}
%In this subsection, we will try to understand the relationship between differential entropy and the PMOG likelihood. 
Suppose $u$ is a random variable with density $P(u)$. The differential entropy of $u$ is given by:

\begin{equation}\label{pmog_diffent1}
H(u) = - \int P(u) \, \log P(u) \, du = - E_u [ \log P(u) ]
\end{equation}
The exact density $P(u)$ of $u$ is not known but suppose we approximate it by a parameterized highly flexible density. It is well known that a mixture of Gaussians with a sufficient number of components $R$ can approximate any probability density with any desired accuracy. Suppose $P^{mog}_R(u \mid \vect{\pi}, \vect{\mu}, \vect{\sigma^2})$ is an $R$-component MOG density as described in \ref{eq4} and \ref{eq4a}. Then we can write:
\begin{equation}\label{pmog_diffent2}
P(u) \approx P^{mog}_R(u \mid \vect{\pi}, \vect{\mu}, \vect{\sigma^2})
\end{equation}
Substituting \ref{pmog_diffent2} in \ref{pmog_diffent1} we see that:
\begin{equation}\label{pmog_diffent3}
H(u) = - E_u [ \log P(u) ] \approx - E_u [ \log P^{mog}_R(u \mid \vect{\pi}, \vect{\mu}, \vect{\sigma^2}) ] 
\end{equation}
Suppose the exact density of $u$ is not known but $n$ independent samples of $u$: $u_1,u_2,\ldots, u_n$ drawn as per $P(u)$ are given. Then we can approximate the expectation in \ref{pmog_diffent3} by a sample average using the law of large numbers. With this approximation we get:
\begin{equation}\label{pmog_diffent4}
H(u) \approx - E_u [ \log P^{mog}_R(u \mid \vect{\pi}, \vect{\mu}, \vect{\sigma^2}) ] \approx - \frac{1}{n} \sum_{i = 1}^n [ \log P^{mog}_R(u_i \mid \vect{\pi}, \vect{\mu}, \vect{\sigma^2}) ]
\end{equation}
Suppose random variable $u$ is related to another random variable $\vect{z}$ via a linear transform $u = \vect{w}^T \vect{z}$ and so the samples $u_i$ are generated by a linear transformation $u_i = \vect{w}^T \vect{z_i}$ then we can write:
\begin{equation}\label{pmog_diffent5}
\boxed{ H(u) = H(\vect{w}^T \vect{z}) \approx - \frac{1}{n} \sum_{i = 1}^n [ \log P^{mog}_R( \vect{w}^T \vect{z_i} \mid \vect{\pi}, \vect{\mu}, \vect{\sigma^2}) ] }
\end{equation}

Suppose our goal is to find a projection vector $\vect{w}$ such that the differential entropy of the projection $\vect{w}^T \vect{z}$ is minimized given realizations of $\vect{z}$: $\vect{z_1},\vect{z_2},\ldots,\vect{z_n}$, Then as shown in equation \ref{pmog_diffent5}, \textbf{minimizing} $H(\vect{w}^T \vect{z})$ is equivalent to \textbf{maximizing} $\sum_{i = 1}^n [ \log P^{mog}_R( \vect{w}^T \vect{z_i} \mid \vect{\pi}, \vect{\mu}, \vect{\sigma^2}) ] $ which is essentially the log likelihood of the PMOG model. In addition to the log likelihood term, the PMOG objective in \ref{eq14} also includes prior terms for $\vect{\pi}$ and $\vect{\sigma^2}$. These prior terms simply prevent the collapse of a PMOG component density onto a single point and thus make the PMOG estimation well conditioned.

\subsection{Sequential source extraction in PMOG based BSS}\label{pmog_sequential}

As discussed in the previous section, the assumption of uncorrelated sources $\matr{\Sigma_s} = \matr{I}_q$ results in the dropping out of the second term in \ref{nonorthogonal5a}. If the sources are correlated, $\matr{\Sigma_s} \neq \matr{I}_q$ then the second term drops out only if $\psi = 0$ i.e., when minimizing differential entropy is considered to be much more important than minimizing correlatedness. The important point to note is that in the uncorrelated source case, a sequential estimation algorithm for BSS using the PMOG model always exists. In the potentially correlated source case, such a sequential algorithm will exist only when $\psi = 0$.

In PMOG based BSS, we simply model the empirical density of $\vect{w_m}^T \vect{z_i}$ by a flexible $R$-component MOG density. As shown in \ref{pmog_diffent5}, this is equivalent to minimizing the differential entropy of the projected points. We can use the PMOG EM algorithm \ref{alg1} to estimate $\vect{w_m}$ along with the MOG parameters under constraints \ref{bss10a} or \ref{nonorthogonal6} depending on whether we want to enforce $\matr{\Sigma_s} = \matr{I}_q$ or $\matr{\Sigma_s} \neq \matr{I}_q$ respectively. Since a MOG  density with sufficiently large $R$ can accurately model any non-Gaussian density, we should be able to extract complex non-Gaussian source densities after solving the PMOG problem. Thus in PMOG based BSS:
\begin{itemize}
\item We replace the variational approximation step in the work of Attias et al.\ \cite{Attias:1999} by an ML step for the PPCA model of Tipping et al.\ \cite{Tipping:1999a} to compute the mixing parameters. This step is similar to that of Beckmann et al.\ \cite{Beckmann:2004}.
\item For latent source estimation, we use a least squares approach after mixing parameters have been estimated. However, we replace the objective functions based on approximation to the differential entropy of Hyvarinen et al.\ \cite{negentropy:1998} with the PMOG model in which the latent sources are assumed to be described by an $R$ component MOG density.
\item PMOG could have a potential advantage in cases where the latent source densities are very complicated. By choosing a sufficiently large $R$ in \ref{pmog_diffent5}, this complicated density can be approximated accurately by the PMOG model. 
\end{itemize}
As discussed above, sequential source estimation is possible under the PMOG model even under partial source dependence (for $\psi = 0$). Thus, the PMOG based BSS retains flexible source density modeling of Attias et al.\ \cite{Attias:1999}, is computationally tractable and can be applied under partial second order dependence between sources.

\section{Experiments and Results}
To illustrate the performance of PMOG based BSS, we performed 2 experiments.
\begin{itemize}
\item In Experiment 1, we generate several artificial "sources" using a MOG model. Next we create mixed data multiple times using the same "sources" but with different random mixing matrices. Note this is a case of "non-square" and "noise free" mixing. For each mixture, we run both FICA and PMOG followed by a statistical comparison of the performance of PMOG with FICA across multiple runs.

\item In Experiment 2, we use real 2-D pictures from a standard image repository as "sources" and mix them using random mixing matrices. Note that this is a case of "square" and "noise free" mixing. We then run FICA and PMOG on the mixed picture data and compare the resulting recovered sources using each method to the "true sources" by visual inspection.
\end{itemize}

In each experiment, we used the implementation of FICA from the FastICA package of Hyvarinen et al.\ (\cite{FPICA:1999},\cite{FPICA:2000}) available from \verb+http://www.cis.hut.fi/projects/ica/fastica/index.shtml+.

\subsection{Experiment1}\label{expt1}
We generated synthetic data with $q = 7$ MOG sources using a mixture of $R=5$ Gaussians. Each MOG source had $n=1000$ sample points. For each MOG source, elements of the class fraction vector $\vect{\pi}$ were chosen from $U(0,1)$ and then normalized to have a sum of 1. Elements of the mean vector $\vect{\mu}$ were chosen from $U(-10,10)$ and the elements of the variance vector $\vect{\sigma^2}$ were chosen from $U(1,5)$. 

Suppose $\matr{S}$ is the $q \times n$ matrix made up of the $n$ realizations of the source vectors $\vect{s_i}$, $\matr{S} = [\vect{s_1},\vect{s_2},\ldots,\vect{s_n}]$. Let $\vect{e_n}$ be a $n \times 1$ vector of all 1's and suppose the eigen-decomposition of the sample co-variance of $\vect{s_i}$ is:
\begin{align}\label{expt1_1}
\vect{\mu_s} &= \frac{1}{n} \sum_{i = 1}^n \vect{s_i} = \frac{1}{n} \matr{S} \vect{e_n} \\
\matr{U_s} \matr{\Lambda_s} \matr{U_s}^T &= \frac{1}{n} \left(\matr{S} - \vect{\mu_s} \vect{e_n}^T\right) \left(\matr{S} - \vect{\mu_s} \vect{e_n}^T\right)^T
\end{align}
Here $\matr{U_s}$ is a $q \times q$ orthogonal matrix and $\matr{\Lambda_s}$ is a diagonal matrix. We construct whitened sources that have $\vect{0}$ sample mean and identity sample co-variance as follows:
\begin{equation}\label{expt1_2}
\matr{\tilde{S}} = \matr{\Lambda_s^{-\frac{1}{2}}} \, \matr{U_s}^T \left( \matr{S} - \vect{\mu_s} \vect{e_n}^T \right)
\end{equation}
The $q \times n$ matrix $\matr{\tilde{S}} = [\vect{\tilde{s_1}}, \vect{\tilde{s_2}}, \ldots, \vect{\tilde{s_n}}]$ holds $n$ samples of the whitened source vectors $\vect{\tilde{s_i}}$. These whitened sources were mixed by random $p \times q$ mixing matrices and embedded into a $p = 20$ dimensional space. We generated a mixture of blind sources $m=50$ times, each time using a different mixing matrix as follows:
\begin{equation}\label{expt1_3}
\matr{X}^{(m)} = \matr{A}^{(m)} \matr{\tilde{S}}
\end{equation}
Here $\matr{A}^{(m)}$ is a random $p \times q$ mixing matrix with elements drawn from $U(0,1)$. The $p \times n$ matrix $\matr{X}^{(m)} = [\vect{x_1^{(m)}},\vect{x_2^{(m)}},\ldots,\vect{x_n^{(m)}}]$ contains the mixed signal vectors in $p$ dimensional space for the $m$th run. We analyzed each $\matr{X}^{(m)}$ as follows:

\begin{itemize}
\item Run FICA on $\matr{X}^{(m)}$ to get $q$ sources as rows of the $q \times n$ matrix $\matr{\tilde{S}_{FICA}^{(m)}}$. We used the default settings for the FICA algorithm from the FastICA package with the odd function $G(x) = \tanh(x)$.
\item Run orthogonal PMOG on $\matr{X}^{(m)}$ to get $q$ sources as rows of the $q \times n$ matrix $\matr{\tilde{S}_{PMOG}^{(m)}}$.\\
\end{itemize}

\begin{defn}\textbf{Match between two matrices}\\
Given two $q \times n$ matrices $\matr{A}$ and $\matr{B}$, let $\rho_{ij}\left(\matr{A}, \matr{B}\right)$ denote the correlation coefficient between the $i$th row of $\matr{A}$ and the $j$th row of $\matr{B}$. We define the "match" between $\matr{A}$ and $\matr{B}$ as a quantity that measures the average value of the best absolute correlation coefficient of the rows of $\matr{A}$ with $\matr{B}$ as follows:
\begin{equation}\label{expt1_4}
\mbox{Match}\left( \matr{A}, \matr{B} \right) = \frac{1}{q} \sum_{i = 1}^q \max_{j} \left[ \mbox{abs} \, \rho_{ij} \left(\matr{A},\matr{B}\right) \right]
\end{equation}
\end{defn}

Our goal is to compare $\mbox{Match}\left( \matr{\tilde{S}}, \matr{\tilde{S}_{FICA}^{(m)}} \right)$ with $\mbox{Match}\left( \matr{\tilde{S}}, \matr{\tilde{S}_{PMOG}^{(m)}} \right)$ across the $m=50$ runs of FICA and PMOG. Since the distribution of the $\mbox{Match}$ values is potentially non-Gaussian, we apply a transformation to Normality to both $\mbox{Match}\left( \matr{\tilde{S}}, \matr{\tilde{S}_{FICA}^{(m)}} \right)$ and $\mbox{Match}\left( \matr{\tilde{S}}, \matr{\tilde{S}_{PMOG}^{(m)}} \right)$ as described in \cite{vanAlbada:2007} followed by a statistical comparison of the transformed values: 
\begin{itemize}
\item In brief, we simply pass the empirical distribution of $\mbox{Match}\left( \matr{\tilde{S}}, \matr{\tilde{S}_{FICA}^{(m)}} \right)$ across $m$ through the inverse CDF of a Normal distribution with the same sample mean and variance. A similar transformation to Normality is applied to $\mbox{Match}\left( \matr{\tilde{S}}, \matr{\tilde{S}_{PMOG}^{(m)}} \right)$ and this transforms the non-Gaussian $\mbox{Match}$ values to Gaussianity. 
\item Let us denote the Normally distributed $\mbox{Match}$ values for FICA and PMOG using the notation $\mbox{Match}_{Normal}\left( \matr{\tilde{S}}, \matr{\tilde{S}_{FICA}^{(m)}} \right)$ and $\mbox{Match}_{Normal}\left( \matr{\tilde{S}}, \matr{\tilde{S}_{PMOG}^{(m)}} \right)$ respectively. We performed a $2$-sample $t$-test with unequal variance to compare the mean of $\mbox{Match}_{Normal}\left( \matr{\tilde{S}}, \matr{\tilde{S}_{FICA}^{(m)}} \right)$ and $\mbox{Match}_{Normal}\left( \matr{\tilde{S}}, \matr{\tilde{S}_{PMOG}^{(m)}} \right)$. 
\end{itemize}

\begin{figure}[htbp]
\begin{center}
\includegraphics[width = 7in] {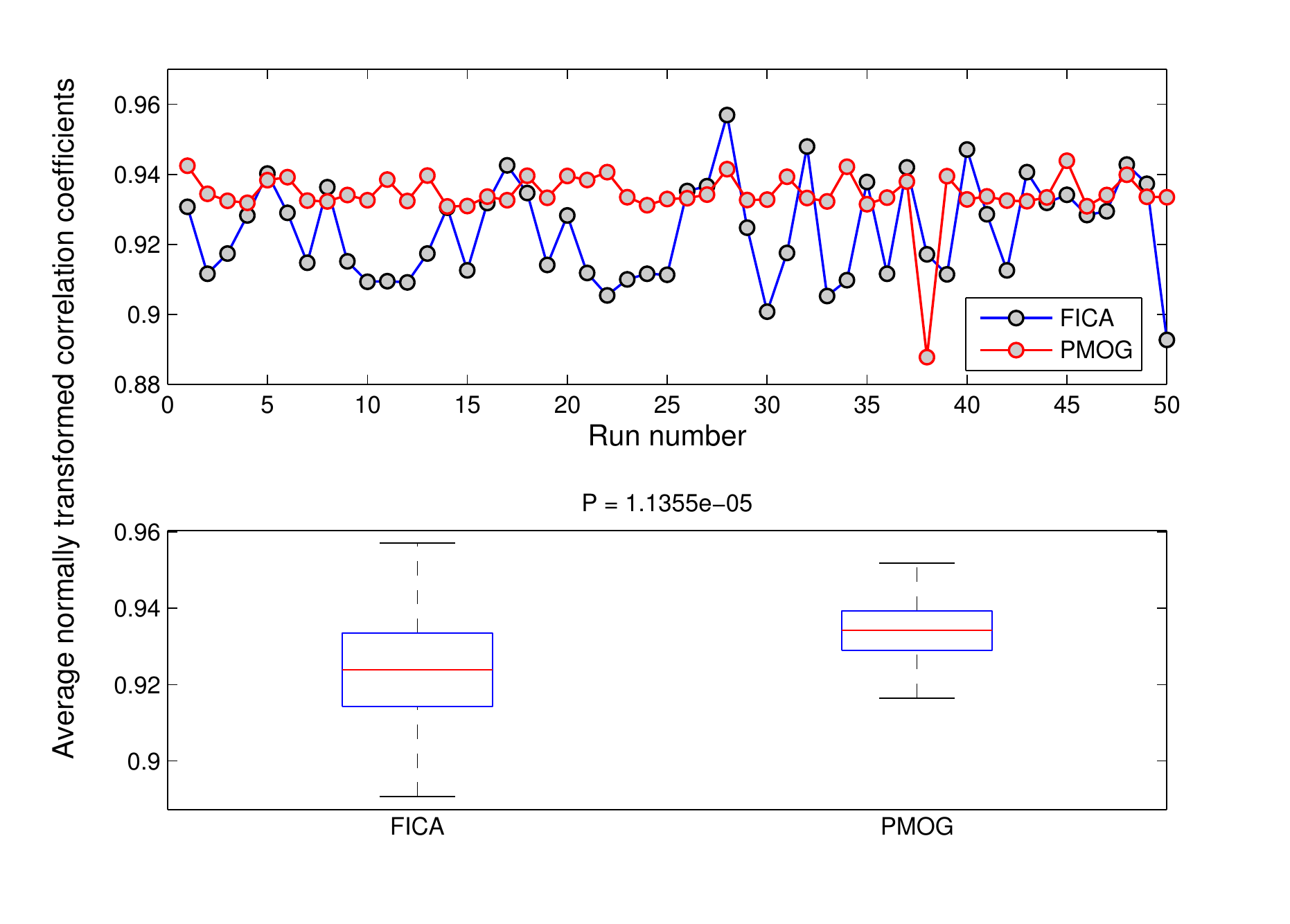}
\caption{The top figure shows the quantities $\mbox{Match}_{Normal}\left( \matr{\tilde{S}}, \matr{\tilde{S}_{FICA}^{(m)}} \right)$ (\textcolor{blue}{blue}) and $\mbox{Match}_{Normal}\left( \matr{\tilde{S}}, \matr{\tilde{S}_{PMOG}^{(m)}} \right)$ (\textcolor{red}{red}) across the $m$ runs. The bottom figure shows the results of a 2-sample $t$-test with unequal variance on the data from the top figure. The results from PMOG are statistically significantly better than those of FICA with a $p$-value of $\sim 1.13 \times 10^{-5}$.}
\label{figure2}
\end{center}
\end{figure}

Results are shown in Fig. \ref{figure2}. PMOG produces results that are statistically significantly better than FICA with a $p$-value of $\sim 1.13 \times 10^{-5}$ across the $m = 50$ runs.

\subsection{Experiment 2}\label{expt2}
To illustrate the performance of PMOG based BSS on real data, we selected 2 separate sets of 3 images from the Berkeley Segmentation Dataset and Benchmark \cite{MartinFTM01}. First, each image was demeaned and standardized to have unit variance. In each case, the 3 identical size images were mixed by a random $3 \times 3$ mixing matrix $\matr{A}$ and mean vector $\vect{\mu}$ whose elements were drawn from a $\mathcal{N}(0,1)$ Normal distribution. Note that this is a case of "square" and "noise free" mixing. In both cases, the 3 mixed images were subjected to 3 different analyses:
\begin{itemize}
\item FICA analysis with the standard orthogonality constraints on the projection vectors. We used the default settings for the FICA algorithm from the FastICA package with the odd function $G(x) = \tanh(x)$.
\item PMOG analysis with orthogonality constraints on the projection vectors.
\item PMOG analysis without orthogonality constraints on the projection vectors.
\end{itemize} 

\begin{figure}[htbp]
\begin{center}
\includegraphics[width = 7in] {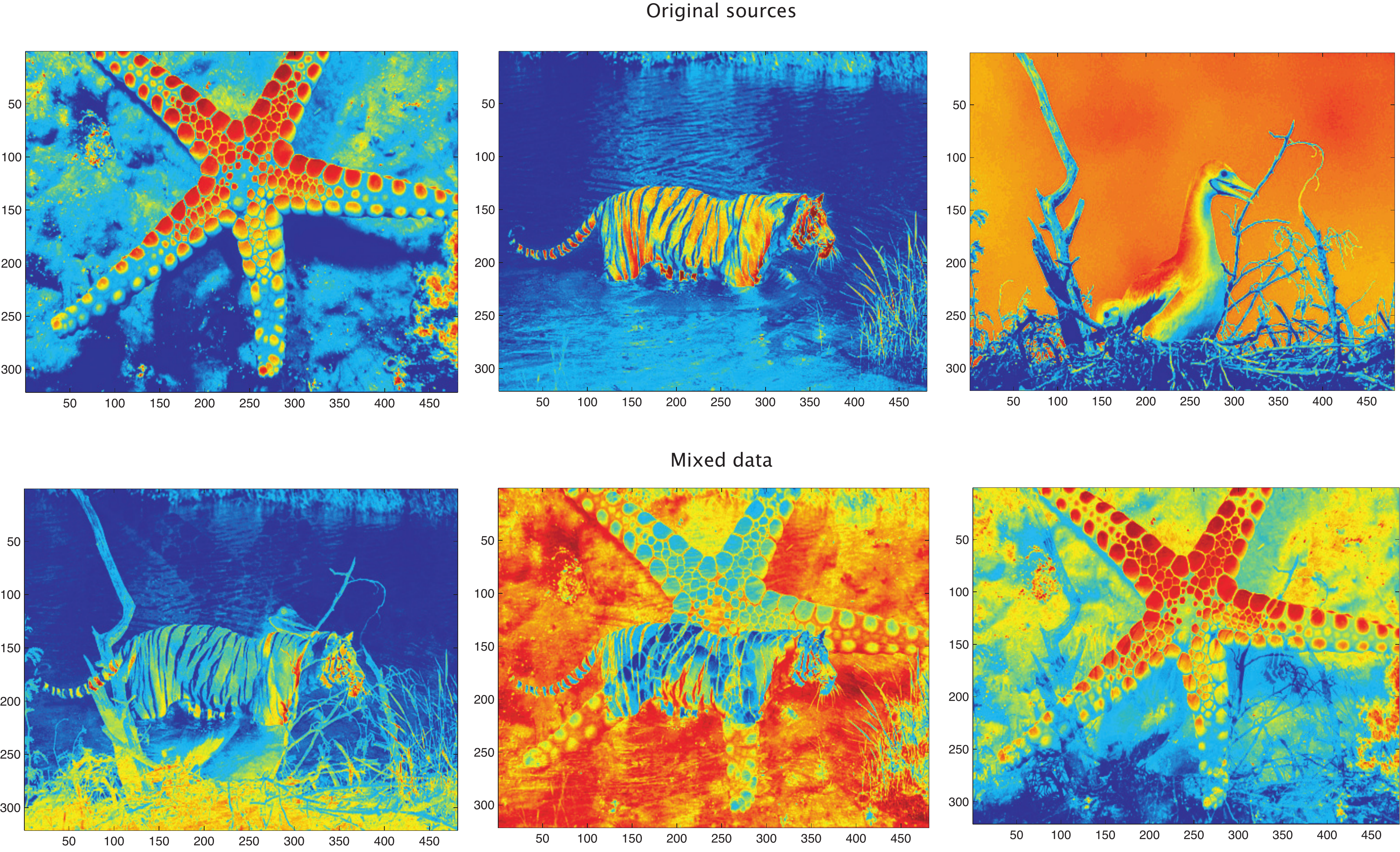}
\caption{ The top row shows 3 pictures of natural images i.e., the original sources. The bottom row shows mixed pictures obtained after mixing image intensities from each pixel of the 3 images using a $3 \times 3$ random mixing matrix and adding a random mean offset. This mixed data was analyzed using FICA and PMOG. Results are shown in Fig. \ref{figure4}.}
\label{figure3}
\end{center}
\end{figure}

\begin{figure}[htbp]
\begin{center}
\includegraphics[width = 7in] {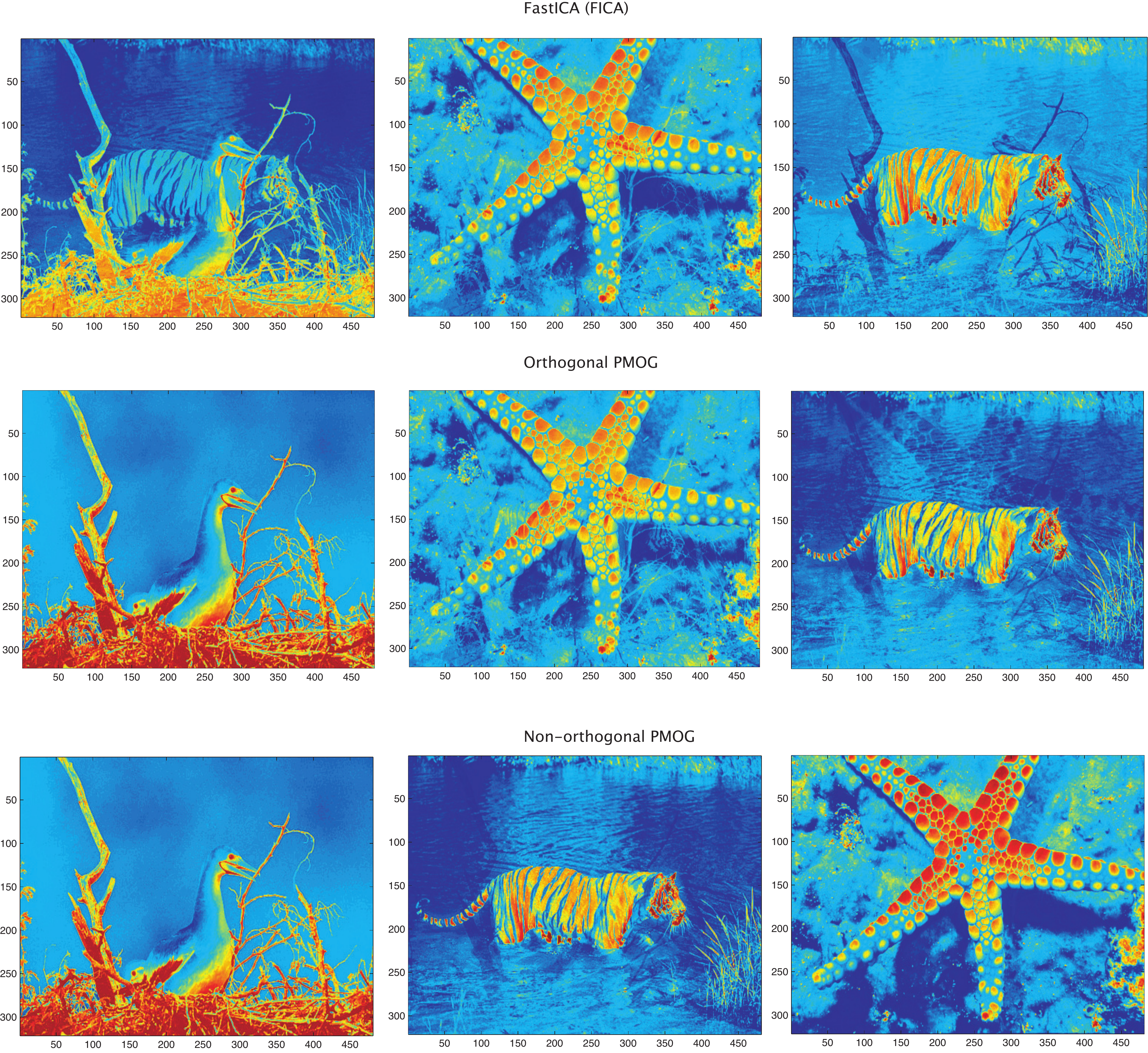}
\caption{ This figure shows the results of running FICA and PMOG on the data from Fig. \ref{figure3}. The top row shows 3 estimated sources using the FICA algorithm. For this dataset the 'defl' approach in FICA fails and hence we used the 'symm' approach. The middle row shows 3 estimated sources using PMOG algorithm with orthogonality constraint on the 3 projections. The bottom row shows estimated sources using the PMOG algorithm without imposing the orthogonality constraint on the 3 projections.}
\label{figure4}
\end{center}
\end{figure}

\begin{figure}[htbp]
\begin{center}
\includegraphics[width = 7in] {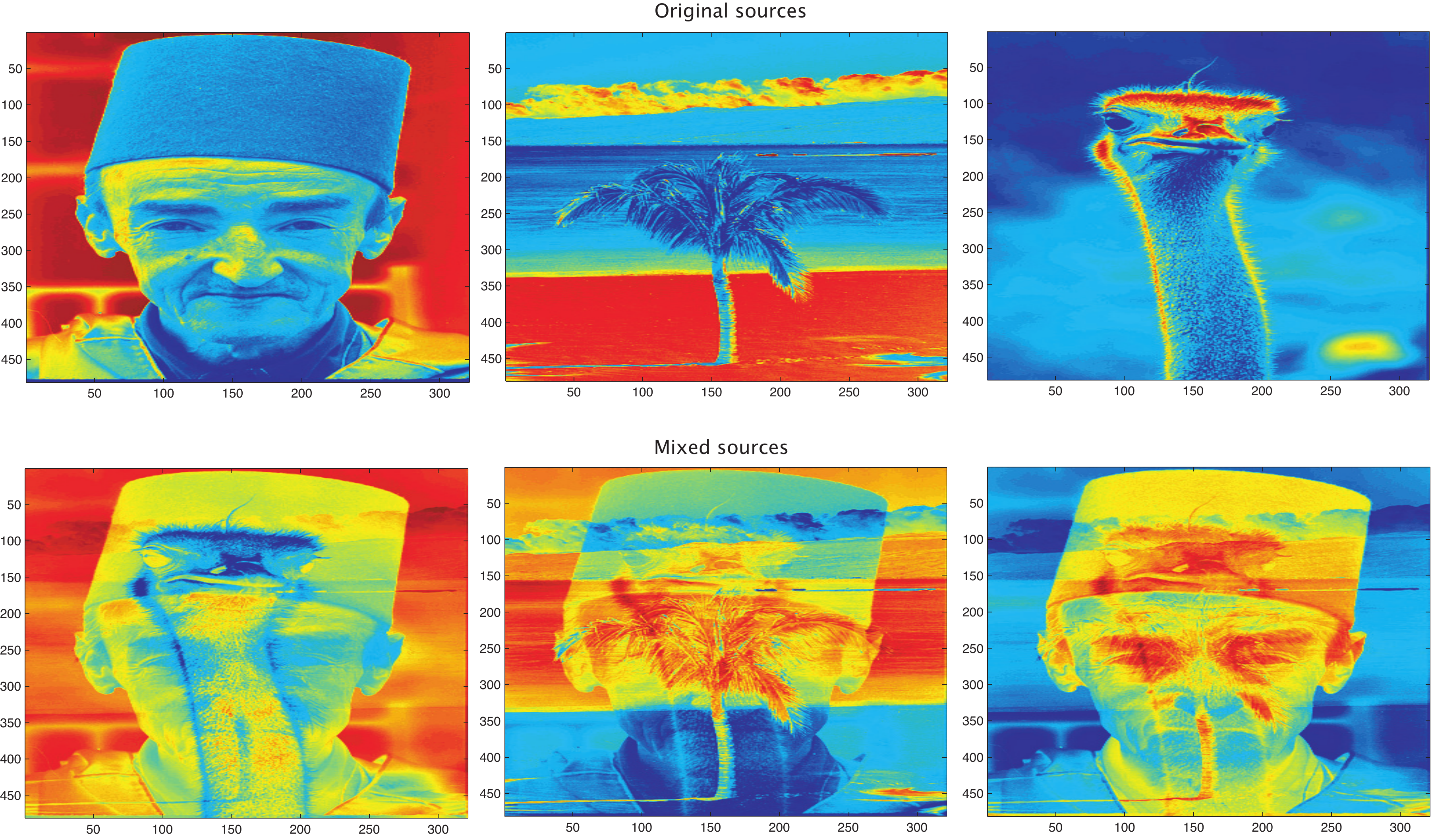}
\caption{ The top row shows 3 pictures of natural images i.e., the original sources. The bottom row shows mixed pictures obtained after mixing image intensities from each pixel of the 3 images using a $3 \times 3$ random mixing matrix and adding a random mean offset. This mixed data was analyzed using FICA and PMOG. Results are shown in Fig. \ref{figure6}.}
\label{figure5}
\end{center}
\end{figure}

\begin{figure}[htbp]
\begin{center}
\includegraphics[width = 7in] {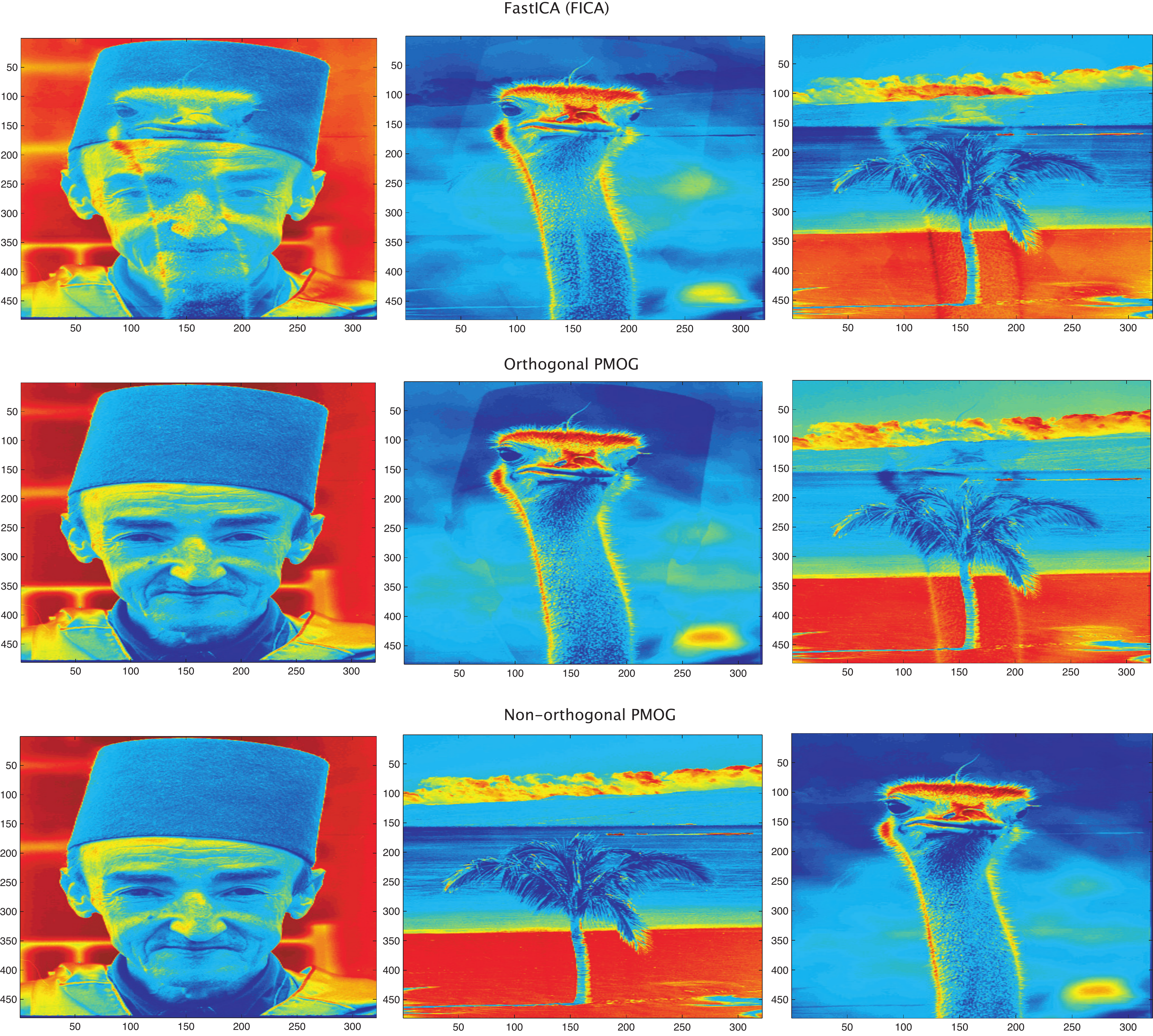}
\caption{ This figure shows the results of running FICA and PMOG on the data from Fig. \ref{figure5}. The top row shows 3 estimated sources using the FICA algorithm. The middle row shows 3 estimated sources using PMOG algorithm with orthogonality constraint on the 3 projections. The bottom row shows estimated sources using the PMOG algorithm without imposing the orthogonality constraint on the 3 projections.}
\label{figure6}
\end{center}
\end{figure}

Results are shown in Fig. \ref{figure3} - \ref{figure6}. For the example shown in Fig. \ref{figure3}, FICA failed to converge using the default settings. Upon experimenting with various settings, we found that the FICA algorithm was able to converge for the data in Fig. \ref{figure3} when we change the "decorrelation" approach to "symm" (see the FICA package for more details). The default settings of FICA resulted in successful convergence for the example in Fig. \ref{figure5}.

It can be seen visually from Fig. \ref{figure4} and Fig. \ref{figure6} that the results of PMOG are better than that of FICA. Since there are dependencies between the intensities of corresponding pixels in the 3 images, they are not exactly independent. Thus, we see that PMOG without orthogonality constraint is able to achieve better source separation compared to PMOG with orthogonality constraint.

\begin{figure}[htbp]
\begin{center}
\includegraphics[width = 5.5in] {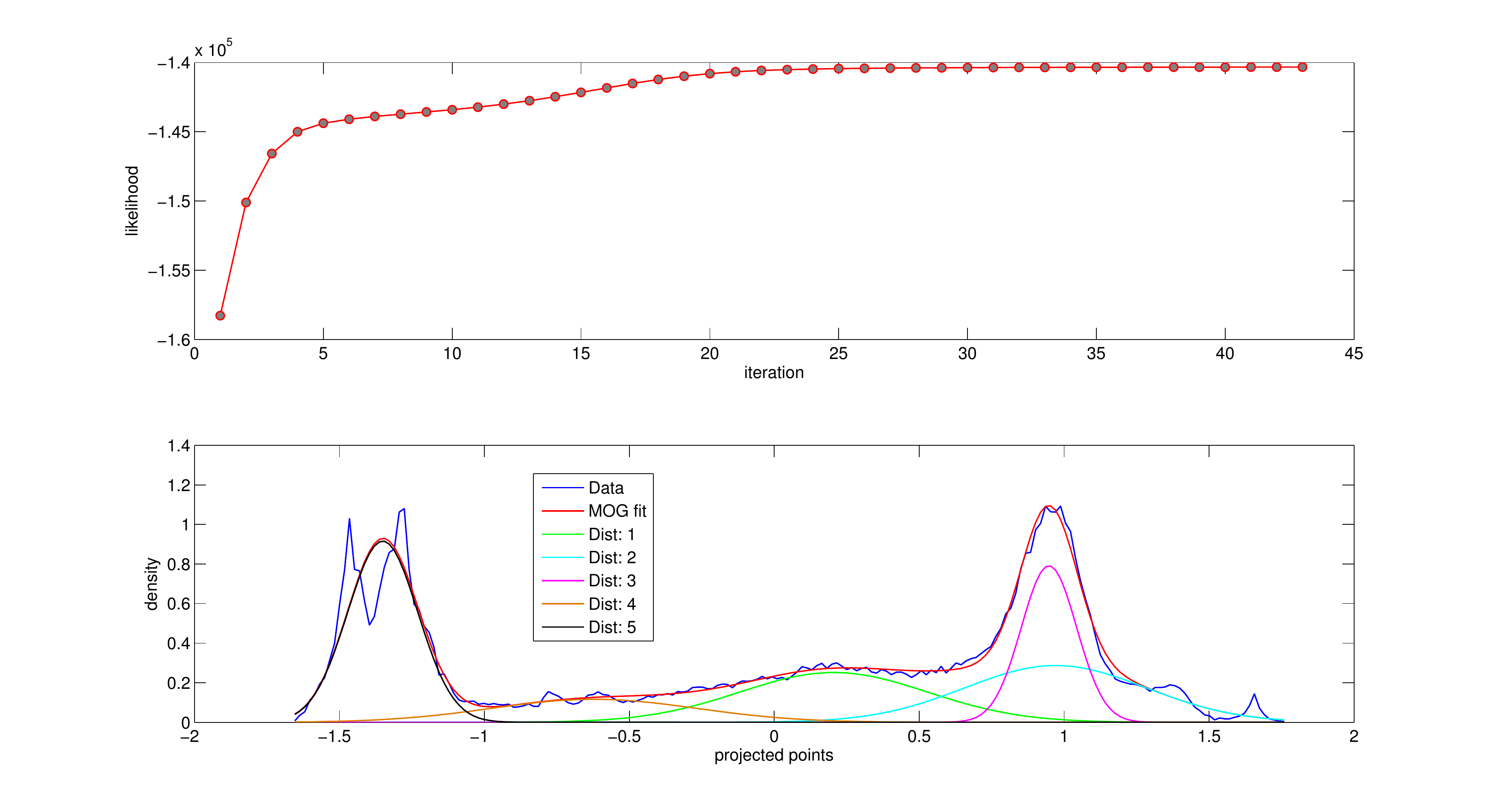}
\caption{ This figure shows the fitted PMOG model for one of the extracted sources (the bottom left image in Fig. \ref{figure6}). Top figure shows the evolution of PMOG log likelihood over iterations illustrating the monotonic increase property of the EM algorithm for estimating the PMOG model. Bottom figure shows the 1-D projected density (corresponding to the estimated projection vector) fitted by a $5$ component PMOG model. In PMOG, both the projection vector and the MOG parameters are jointly optimized.}
\label{pmog_illustration}
\end{center}
\end{figure}

As an example, we also show the fitted PMOG model for one of the extracted sources (the bottom left image in Fig. \ref{figure6}) in Fig. \ref{pmog_illustration}. This figure shows the monotonic increase in PMOG objective function $H(\vect{\pi}^{(t)}, \vect{\mu}^{(t)}, \vect{\sigma^2}^{(t)}, \vect{w}^{(t)})$ from \ref{eq22} over EM iterations $t$ and the final fitted PMOG model using $R = 5$ Gaussian distributions. It is interesting to note that this distribution is highly non-Gaussian and hence approximations to negentropy used in FICA based on the "near Gaussianity" assumption in \cite{negentropy:1998_techrep} might not be adequate in such cases.

\section{Discussion}\label{discussion}

In this work, we posed the problem of estimating a projection of input data such that the projection is well described by an $R$ component MOG density. We showed that it is possible to derive an EM algorithm for solving this problem. Since the estimation of projection vector is coupled with the estimation of distributional parameters, we break up the M-step into two parts:
\begin{itemize}
\item In part 1 of the M-step, we estimate the distributional parameters for a fixed projection vector. 
\item In part 2 of the M-step, we optimize for the projection vector while fixing the distributional parameters. 
\end{itemize}
We show that solving the M-step problem in part 2 is equivalent to finding the roots of a particular cubic equation for the projection vector. Since the objective in part 2 of the M-step is non-concave, the solution to the set of cubic equation in part 2 might converge to a minimum instead of a maximum. To enforce the monotonic likelihood increase property of the EM algorithm, we explicitly check equation \ref{eq23a} after each M-step. If this condition is not satisfied, we randomly re-initialize $\vect{w}$ and repeat the M-step.

\begin{figure}[htbp]
\begin{center}
\includegraphics[width = 6.5in] {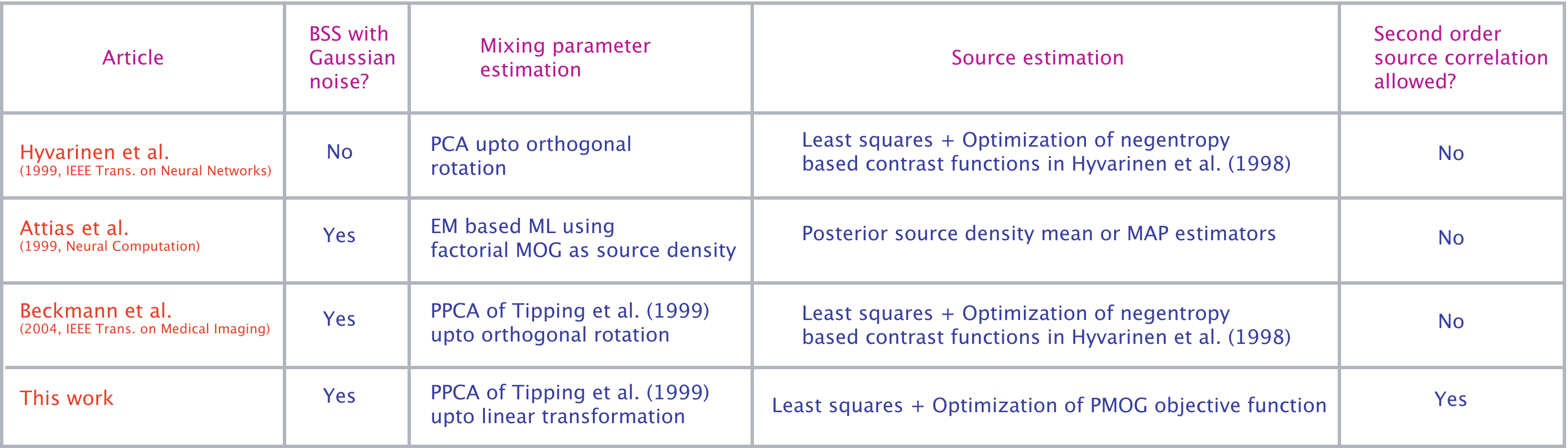}
\caption{A comparison of the features and estimation techniques of four different BSS algorithms.}
\label{various_algorithms}
\end{center}
\end{figure}

Next we considered the application of the PMOG model to the solution of the BSS problem. If the mixing process is without Gaussian noise then the estimation of mixing parameters and latent sources is uncoupled \cite{Hyvarinen:1999}. In this case, the BSS problem can be reduced to one of estimating an orthogonal matrix by using principal component analysis (PCA) as a preprocessing step. However, in the presence of additive Gaussian noise, the estimation of mixing parameters and latent sources is coupled and this makes the BSS problem a difficult one. 

Attias et al.\ \cite{Attias:1999} developed a very general solution for the BSS problem under the assumption of exact independence between latent sources. In particular, \cite{Attias:1999} assumed an MOG density for each latent source. This implies a "factorial MOG" joint density for the vector of latent sources under independence. Further, \cite{Attias:1999} derived an EM algorithm for joint estimation of both the mixing parameters and "factorial MOG" parameters. While this work provides an exact solution to the BSS problem, it assumes exact independence between sources and becomes computationally intractable for $> 13$ latent sources. As a computationally tractable alternative, \cite{Attias:1999} propose a variational approximation to compute the BSS parameters followed by using a MAP or posterior mean estimate for the sources.

Beckmann et al.\ \cite{Beckmann:2004} proposed a BSS solution in which the estimation of mixing parameters and latent sources is de-coupled by assuming that the sources have joint distribution $\mathcal{N}(\matr{0},\matr{I}_q)$. This effectively reduces the BSS model to the PPCA model of Tipping et al.\ \cite{Tipping:1999a}. In \cite{Beckmann:2004}, the mixing parameters are chosen to be the ML solution for this PPCA model after which the BSS problem reduces to one of estimating an orthogonal matrix. In both \cite{Hyvarinen:1999} and \cite{Beckmann:2004}, the sources are estimated as orthogonal projections that minimize the sum of differential entropies of the projections as in equation \ref{bss10}. Both techniques \cite{Hyvarinen:1999} and \cite{Beckmann:2004} use approximations to the differential entropy developed in \cite{negentropy:1998} based on the assumption of "near Gaussian" source densities. 

While the solution by Attias et al.\ \cite{Attias:1999} is very attractive (at least under exact independence) since it allows flexible source density modeling via a "factorial MOG", it suffers from computational intractability for $>13$ sources. Our work bridges the gap between the approach of Attias et al.\ \cite{Attias:1999} and that of Hyvarinen et al.\ \cite{Hyvarinen:1999} and Beckmann et al.\ \cite{Beckmann:2004}. On the one hand, it retains computational tractability by using the PPCA approach of Beckmann et al.\ and on the other hand it allows for flexible source density modeling of Attias et al.\ via the PMOG model. As we show in \ref{pmog_diffent5}, minimizing the differential entropy is equivalent to maximizing the PMOG likelihood function. In the PMOG model, we jointly estimate both the projection vector as well as MOG distributional parameters to maximize the likelihood of observing an MOG in the projected space. The overall algorithm retains computational feasibility for sources $>13$ while retaining the flexibility of modeling arbitrarily complex source densities (given enough Gaussians in PMOG). 

In the work of Attias et al.\ \cite{Attias:1999} an approximation is made to true ML objective (based on "factorial MOG" sources) using variational arguments to retain computational tractability for $> 13$ sources. In the present work and in Beckmann et al.\ \cite{Beckmann:2004}, exact ML solution is used for the analytically tractable PPCA model (based on $\mathcal{N}(\matr{0},\matr{I}_q)$ sources). Both methods use approximations. Is the variational approximation more accurate than the PPCA approximation for BSS parameters? This question is open for discussion and beyond the scope of this work. However, based on our experiments with real and synthetic data the PPCA approximation followed by application of the PMOG results in good performance. We also show that when the latent sources are only approximately independent, then we should simply run the PMOG algorithm with only the unit norm constraints on the projection vectors (i.e. without the orthogonality constraints). Thus PMOG based BSS generalizes elegantly to cases where non-zero second order correlation exists between sources. Fig. \ref{various_algorithms} shows a comparison of PMOG based BSS with alternative approaches.

It is worth noting the difference between the BSS algorithms in Fig. \ref{various_algorithms} and a related technique from statistics - projection pursuit (PP) \cite{Friedman:1974, Huber:1985, Friedman:1987}. In BSS, the problem is to estimate both the mixing parameters and latent sources given the linear mixing model \ref{bss1}. As shown in \cite{Attias:1999}, this is a coupled estimation problem. However, if we use a PPCA step to estimate mixing parameters then this problem becomes decoupled and the latent sources can be estimated using least squares upto a linear transformation. If sources are assumed to be uncorrelated then this linear transformation is orthogonal. At this point, the problem becomes similar to a PP problem, where linear transformations (not necessarily orthogonal) are sought that optimize general contrast functions. Whereas in PP any contrast function can be used, the focus in BSS is to maximize the independence between sources and so a linear transformation that minimizes the mutual information between the sources is sought. Thus PMOG based BSS can also be thought of as a combination of PPCA preprocessing followed by PP using the special PMOG objective function which is optimized using an EM algorithm with or without orthogonality conditions on the projections.

Both FICA and PMOG solve an optimization problem where different objective functions are maximized. In the case of FICA, the objective function is an approximation to differential entropy developed in \cite{negentropy:1998} whereas in the case of PMOG the objective function is the PMOG log likelihood \ref{pmog_diffent5}. In both FICA and PMOG, the objective functions are not concave and so the solutions to the maximization problems are not unique. An ideal solution would be to find the global solution to the maximization problems in FICA and PMOG and compare the results. Since finding the global solution itself is a difficult problem an alternative suboptimal method of comparing FICA and PMOG is to run each algorithm multiple times on mixtures generated using the same "sources" and then compare the quality of results across runs. This is the approach we used in Experiment 1 \ref{expt1}. It was found that on average PMOG produced significantly better results compared to FICA across runs. We would like to mention that since the true "sources" in Experiment 1 are in fact MOG sources, the results could be slightly biased towards PMOG. Nevertheless, Experiment 1 does illustrate that when the true "sources" have a complicated density (such as a MOG) then PMOG might show improved performance.

We also ran 2 illustrative experiments on publicly available natural image datasets \cite{MartinFTM01} in Experiment 2 \ref{expt2} and compared the results of FICA and PMOG by visual inspection. FICA showed relatively poor performance compared to orthogonal PMOG. We postulate that this is because the source densities are multimodal (see Fig. \ref{pmog_illustration}) and so the PMOG model captures them more faithfully compared to the approximate cost function \cite{negentropy:1998} used in FICA. We also noticed that assumption of orthogonality of projection vectors also hurts the separation performance because the source images have non-zero second order correlation and hence are not exactly independent. The best performing algorithm was non-orthogonal PMOG in which maximal source independence is enforced with potentially non-zero second order correlation (i.e., a minimal level of dependence is allowed). 

A limitation of PMOG based BSS is the running time of the PMOG algorithm. It is clear from numerical experiments that PMOG is much slower than FICA. For the experiments presented in this work, PMOG takes upto 7 minutes per source estimation versus only a fraction of a second for FICA using a computer with $2 \times 2.93$ GHz Quad-Core Intel Xeon processor with 16 GB RAM. Another drawback is the potential re-initialization required to enforce "non-orthogonality" in the form of PMOG where partial source dependence is allowed. A natural solution to this problem is to perform a joint PMOG estimation with $\psi \neq 0$ in \ref{nonorthogonal5a} which would automatically encourage orthogonality while not explicitly enforcing it.

Finally, note that the approximation in \ref{pmog_diffent5} becomes increasingly accurate as $n$ increases and thus ideally we would like the number of available data samples $n$ to be large. However, this dependence of the quality of approximation on $n$ is not unique to PMOG based BSS. For instance, even in the differential entropy approximations of Hyvarinen et al.\ \cite{negentropy:1998} such as \ref{maxent5}, the expectations on the right hand side are approximated by sample averages which also become increasingly accurate as $n$ increases.

\section{Conclusions}\label{conclusion}
We propose that the non-square linear BSS model with Gaussian noise can be estimated by first using the PPCA model of Tipping et al.\ \cite{Tipping:1999a} followed by application of the PMOG algorithm. The MOG density provides a flexible model for the unknown source density and simulations on illustrative data sets indicate that this approach might be a useful alternative to well established approaches such FICA. Furthermore, it is possible to allow for non-zero second order correlation between latent sources simply by relaxation of the orthogonality requirement in PMOG. This could result in better performance since the assumption of exact statistical independence is unlikely to be true for real world data-sets.

The current version of PMOG based BSS solves the optimization problem in \ref{nonorthogonal5a} for $\psi = 0$ since this decouples the estimation of individual projection vectors. Future work would involve the joint solution of all projection vectors using a PMOG approach for $\psi = 1$.

\section*{Acknowledgements}
We gratefully acknowledge financial support from the Pain and Analgesia Imaging and Neuroscience (P.A.I.N) group, McLean Hospital, Harvard Medical School, Belmont MA, USA under a grant from the Louis Herlands Fund for Pain Research (DB and LB).

\newpage
\bibliographystyle{plain}
\bibliography{bibliography_pmog.bib}

\end{document}